\documentclass[10pt,twocolumn,letterpaper]{article} % DO NOT CHANGE THIS
\pdfoutput=1
\usepackage{times}  % DO NOT CHANGE THIS
\usepackage{helvet}  % DO NOT CHANGE THIS
\usepackage{courier}  % DO NOT CHANGE THIS
\usepackage{graphicx} % DO NOT CHANGE THIS
\usepackage{caption} % DO NOT CHANGE THIS AND DO NOT ADD ANY OPTIONS TO IT
\usepackage{amssymb}
\usepackage{amsmath}

\usepackage{color}
\usepackage{algorithm}
\usepackage{algorithmic}

\usepackage{newfloat}
\usepackage{listings}
\usepackage[colorlinks,linkcolor=blue]{hyperref}
\usepackage{authblk}

\usepackage{bibentry}
\begin{document}
	
\title{Expression Snippet Transformer for Robust Video-based Facial Expression Recognition}
\author[1]{Yuanyuan Liu}
\author[1]{Wenbin Wang}
\author[1]{Chuanxu Feng}
\author[1]{Haoyu Zhang}
\author[2]{Zhe Chen\footnote{Corresponding author}}
\author[3]{Yibing Zhan}
\affil[1]{China University of Geosciences (Wuhan)}
\affil[1]{\textit{\{liuyy, wangwenbin, fcxfcx, zhanghaoyu\}@cug.edu.cn}}
\affil[2]{The University of Sydney}
\affil[2]{\textit{zhe.chen1@sydney.edu.au}}
\affil[3]{Jingdong}
\affil[3]{\textit{zhanyibing@jd.com}}

\maketitle
%%%%%%%%% ABSTRACT
\begin{abstract}
The recent success of Transformer has provided a new direction to various visual understanding tasks, including video-based facial expression recognition (FER). By modeling visual relations effectively, Transformer has shown its power for describing complicated patterns. However, Transformer still performs unsatisfactorily to notice subtle facial expression movements, because the expression movements of many videos can be too small to extract meaningful spatial-temporal relations and achieve robust performance. To this end, we propose to decompose each video into a series of expression snippets, each of which contains a small number of facial movements, and attempt to augment the Transformer's ability for modeling intra-snippet and inter-snippet visual relations, respectively, obtaining the Expression snippet Transformer (EST). In particular, for intra-snippet modeling, we devise an attention-augmented snippet feature extractor (AA-SFE) to enhance the encoding of subtle facial movements of each snippet by gradually attending to more salient information. In addition, for inter-snippet modeling, we introduce a shuffled snippet order prediction (SSOP) head and a corresponding loss to improve the modeling of subtle motion changes across subsequent snippets by training the Transformer to identify shuffled snippet orders. Extensive experiments on four challenging datasets (\textit{i.e.}, BU-3DFE, MMI, AFEW, and DFEW) demonstrate that our EST is superior to other CNN-based methods, obtaining state-of-the-art performance.Our code and the trained model are available at \href{https://anonymous.4open.science/r/ATSE-C58B}{https://anonymous.4open.science/r/ATSE-C58B}

\end{abstract}

%%%%%%%%% BODY TEXT
\section{Introduction}
\begin{figure}[!t]
\begin{center}
	\includegraphics[width=1.0\linewidth]{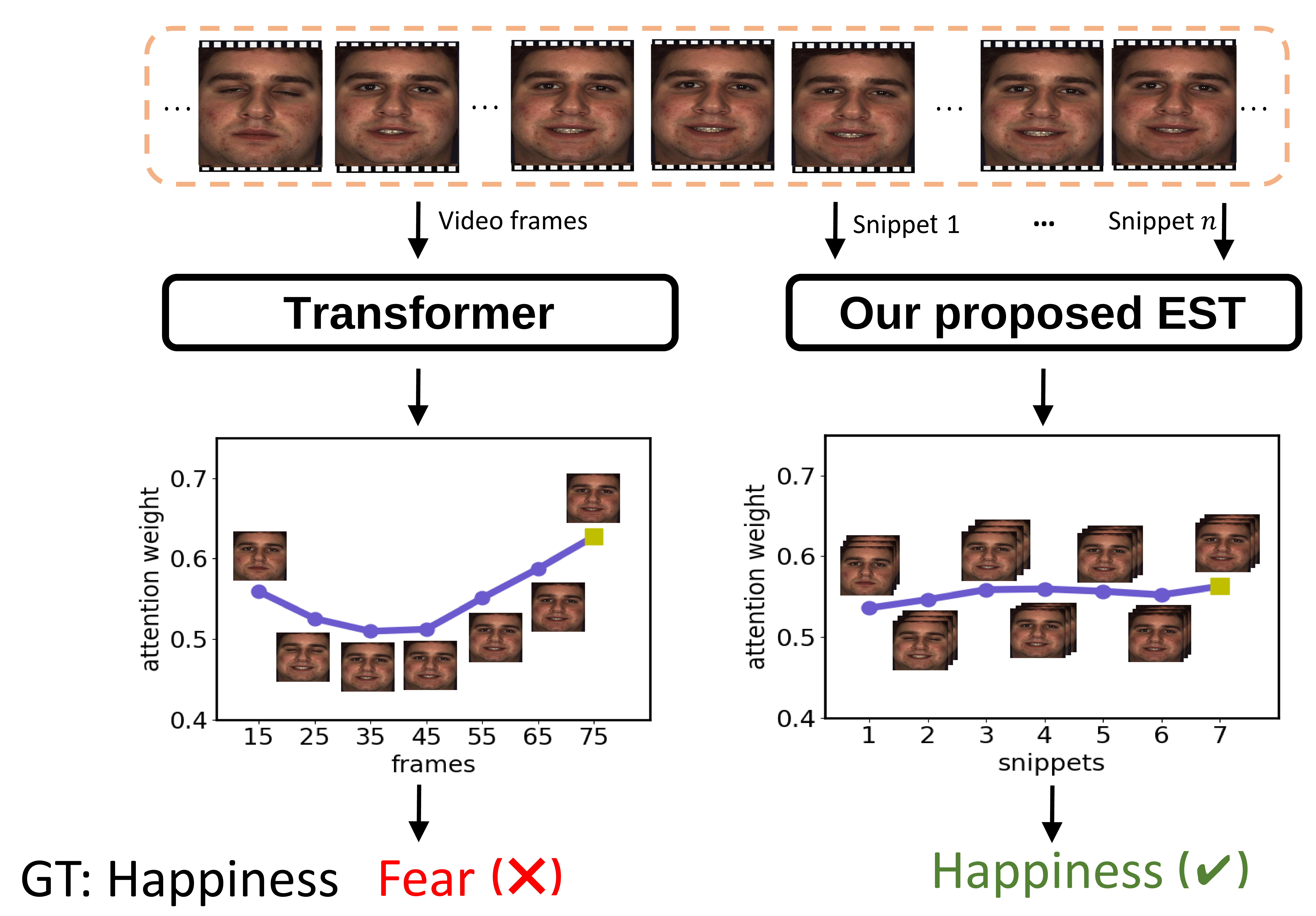}
\end{center}
\vspace{-0.3cm}
	\caption{
	{ Comparison between a vanilla Transformer and the proposed expression snippet Transformer (EST) for modeling subtle facial expression movements in facial expression recognition (FER). The vanilla Transformer (left) tends to focus only on the frame with peak expression patterns and can be easily affected by noises such as other non-expression changes, thus obtaining suboptimal results.
    By decomposing videos into snippets, the EST (right) improves the modeling of intra-snippet and inter-snippet subtle facial changes, respectively, and can achieve more robust FER. }
%can be easily affected by noises in the video. 
	%{\color{red}A comparison between modeling subtle facial expression movements using Transformer and the proposed expression snippet Transformer (EST). Compared with using the Transformer to pay more attention to peak expression in a video, the EST can augment the Transformer's ability for modeling subtle expression movement relations of the entire video and achieving robust expression recognition.} %In the EST, we decompose the video into snippets. Then, we introduce the attention-augmented snippet feature extractor (AA-SFE) and the shuffled snippet order prediction (SSOP) to boost the Transformer for modeling intra-/inter-snippet subtle expression movements, respectively.
	%to boost the modeling of inter-snippet motion relations. 
	%the bottom figure shows the procedure of  with the transformer for modelling inter-snippet motion relations via the snippet features. Note that the expression motion changes in both intra-/inter snippet are small and subtle in video FER. %  We tend to decompose the input video into several expression snippets and obtain more salient expression representation by applying Transformer to model snippet relations and facial motion changes. We propose a novel attention-augmented snippet feature extractor (AA-SFE) to improve the modeling of intra-snippet visual changes, and then devise a shuffled snippet order prediction (SSOP) head to help train the Transformer for modeling inter-snippet relations.
	}
\label{fig:motivation}
\vspace{-0.5cm}
\end{figure}

Video-based Facial Expression Recognition (FER) is important for understanding human emotions and behaviors in videos, benefiting various applications such as digital entertainment, customer service, driver monitoring, emotion robots, etc.~\cite{zhang2016deep,tawari2013face,wu2017locality,lee2019context}. FER aims to classify a video into one of several basic emotions, including happiness, anger, disgust, fear, sadness, neutral, and surprise. 
The task of FER is difficult due to several challenges, namely, long-range spatial-temporal representation, excessive noises from irrelevant frames, and especially, inherently \emph{small} and \emph{subtle} facial movements in FER videos. %Automatically Video-based FER is crucial to They easily cause two problems in FER, that is, it is difficult to capture salient emotion representation and difficult to distinguish motion variation of different expression categories. %intra-class variation and inter-class similarity. %Hence, we hope to propose a robust FER model that can simultaneously predict subtle expression change and discovering the positions of the most emotion-rich representation.

%To better tackle these issues, we propose a robust Transformer-based FER model to simultaneously discover subtle facial movement change and tackle the FER task more effectively. We first decompose the expression movements within a video into a series of short snippets. Then, we incorporate the outstanding relation modelling ability of Transformer \cite{ISI:000452649406008} to help obtain vastly advanced representation of facial expressions for efficacious FER. To help the Transformer gain the ability of depicting subtle facial changes appropriately, we further introduce an attention-augmented snippet feature extractor and a snippet order descriptor for improving the modeling of intra- and inter-snippet visual changes with the Transformer, respectively. } We name the proposed method as expression snippet Transformer (EST). 
%Different from  to tackle these issues, we can also take advantage of the Transformer
%we propose a novel Transformer-based method to address these problems, thus achieve robust FER. 

To tackle the issues of FER, existing methods commonly apply convolutional neural networks (CNNs) or long-short term memory (LSTM).
However, most of the existing FER methods usually model spatial-temporal visual information without involving effective visual relation reasoning mechanisms. For example, many methods \cite{huang2010new, wang2012facial, lee2016collaborative, yang2018facial, kim2017deep} only use static frames selected from the manually defined peak (apex) frames, neglecting the intrinsic relationships between visual cues of adjacent frames. Sequence-based methods \cite{fan2016video, fan2018video, bargal2016emotion, chen2018less} attempt to capture motion cues by encoding spatial-temporal information within their models. However, Sequence-based methods still perform weakly in describing subtle expression movements in FER videos. Besides, they usually require overwhelmingly large model capacities to help model subtle facial changes \cite{li2020deep}. %实际上我对这个问题是存疑的，大模型容量，除非你的方法解决了这个问题
%One of the major drawbacks of these methods is that they introduce overwhelmingly large models and expensive computations to {\cmt better} model {\cmt subtle expression movements in videos implicitly} {\chg for} more robust facial recognition results. 
%On the contrary, we observe that the facial expression movement changes in a video is very weaker than other actions, not only in facial local spatial movements but also in emotion intensity temporal changes. Modelling these emotion changes for effective spatial-temporal representation in a video is very difficult.
%we observe that there is a large amount of frames that only deliver trivial information for FER in a video. Modelling these frames exhaustively can waste the introduced large model capacities. 

The recent successful Transformer approaches \cite{devlin2018bert,luscher2019rwth,synnaeve2019end} in computer vision has allowed us to take advantage of its powerful relation reasoning ability for understanding FER videos. 
%, we propose that modeling relations of different facial change moments with Transformer can be beneficial to depict subtle expression movements and achieve robust FER. 
In general, the Transformer \cite{ISI:000452649406008} has been shown to be particularly effective for translating an input sequence to a target sequence by modeling the relations between features. 
Accordingly, for video-based FER, we suggest that the Transformer has a great potential of describing subtle expression movements more robustly.
%by applying it to model the relations between visual contents from different moments from the input video. 
However, despite the potential advantages, it is non-trivial to directly apply the vanilla Transformer on the FER videos, considering the subtle facial expression movements of videos that are difficult to be noticed by vanilla Transformer.
%In the task of video-based FER, the Transformer can also be helpful for describing the subtle expression movement changes by modeling relations between visual changes of different moments in the video. The modeled relations can facilitate the overall FER model to better focus on meaningful expression information, thus obtaining more appropriate expression representation for FER. Despite the potential advantages, we find that it is still difficult to straightly apply the vanilla Transformer on the FER videos. 
%More specifically, we find that the subtle visual changes between frames could make the Transformer quite difficult to focus on meaningful visual cues.
For example, as shown in Fig.~\ref{fig:motivation}, the per-frame visual information, raw pixels on each frame, may contain noises such as other non-expression changes (head poses, speaking, and so on) that can easily affect the recognition performance of the Transformer. Furthermore, the subtle expression movements would make the Transformer only focus on the visual cues from frames with peak expression changes and neglect plenty of beneficial spatial-temporal information from other periods of videos. This limits the potential of Transformer to encode the motion information of the entire video comprehensively and achieve more robust expression recognition.

To tackle the above problems for applying Transformer on FER videos, we first propose to decompose the facial movements of the entire video into a series of small expression snippets. Each expression snippet is a video clip with a few adjacent frames of the input video covering a limited amount of expression changes. Then, by employing the Transformer over the snippets, we can augment the modeling of intra-snippet and inter-snippet expression movements, respectively. In particular, we introduce a novel attention-augmented snippet feature extractor (AA-SFE) to improve the modeling of intra-snippet visual changes for the Transformer. In the AA-SFE, we apply a deep convolutional neural network (DCNN) to extract per-frame visual features and develop a novel hierarchical attention-augmentation architecture to obtain the representation of facial movements within each snippet. 
%这里可能还需要指出获取的是snippet的feature,所以我改成了obtain
The snippet representations generated with the AA-SFE are subsequently fed into the encoder-decoder structure of the Transformer to perform recognition based on snippet-level relations.  Meanwhile, 
we devise a shuffled snippet order prediction (SSOP) head with a corrsponding loss for the Transformer to improve the modeling of inter-snippet visual changes. % Meanwhile, we devise a shuffled snippet order prediction (SSOP) head with corresponding loss for the Transformer to improve the modeling of inter-snippet visual changes. 
%to pay attention to all the available visual information comprehensively for understanding correct visual change order. 
By using SSOP, the Transformer can encode the information from all snippets more comprehensively, thereby delivering a more robust expression movement representation of the entire video. Overall, we briefly name our proposed method as expression snippet Transformer (EST).

To sum up, the major contributions of this paper are summarized as follows:
\begin{itemize}
	\item We propose the expression snippet Transformer (EST) to achieve accurate video-based facial expression recognition (FER). To the best of our knowledge, our approach is the first effective snippet-based Transformer method for video-based FER.  
	\item To enhance the Transformer's ability to model intra-snippet and inter-snippet expression movements, we propose the attention-augmented snippet feature extractor (AA-SFE) and the shuffled snippet order prediction (SSOP), respectively. Both techniques effectively tackle the problems of Transformer-based FER and substantially improves the recognition performance. 
	
	%enable Transformer to gain the ability of depicting subtle facial changes appropriately in intra-/inter snippets.
	
	%tackle the issues of snippet feature extraction and visual change prediction, we introduce attention-augmented snippet feature extractor (AA-SFE) and a snippet order descriptor (SOD), respectively. Both techniques effectively enable Transformer to gain the ability of depicting subtle facial changes appropriately in intra-/inter snippets. 
	%The first Transformer-based FER method which delivers efficient and effective video-based FER via two mutually-reinforced learning components, namely CFE and TEAT. A new learning scheme-based TEAT is introduced to learn emotion activation mappings from clip-based features extracted by CFE. 
	%an attention assigned multi-head self-attention is used for modeling long-range temporal dependence on video sequence. 
	\item Evaluations on four challenging video facial expression datasets, \textit{i.e.}, BU-3DFE, MMI, AFEW, and DFEW, demonstrate the superiority of our proposed EST over existing popular methods. State-of-the-art performance can be achieved with EST on the evaluated datasets. 
	%\item  A TEAT with temporal deconstruction-reconstruction learning is designed for helping the Transformer identity and enhance the long-temporal relationship of inter-clips in untrimmed videos, improving the learning capacity of the Transformer and preventing the loss of video-level temporal information.
	%\item A CFE is used to transform the temporal visual data into the data space that can be easily accepted by the Transformer, which fully explore spatial emotion representation within clips via jointly learning inter-frame and extra-frame attentions.
%	\item Training code and pre-trained models will be made publicly available upon publication.
\end{itemize}

\section{Related Work}\label{sec:Related}
%In this section, we discuss the methods related to the proposed EST. We mainly review the two most important categories of methods, \textit{i.e.}, static frame-based and dynamic sequence-based, for video-based FER~~\cite{li2020deep}. Typical Transformer methods in language processing and vision are also reviewed.
%Among recent existing works, static frame-based methods and dynamic sequence-based methods are the two most important categories of methods for video-based FER~~\cite{li2020deep}.

\textbf{Frame-based methods} The frame-based methods can be divided into two groups: frame aggregation methods that strategically fuse deep features learned from static-based FER networks~\cite{meng2019frame,knyazev2017convolutional} and peak frame extraction methods that focus on recognizing the peak high-intensity expression frame~\cite{zhao2016peak,yu2018deeper}. Meng \textit{et al.}~\cite{meng2019frame} proposed frame attention networks to adaptively aggregate frame features in an end-to-end framework and achieved an accuracy of 51.18\% on the AFEW 8.0 dataset. Zhao \textit{et al.}~\cite{zhao2016peak} proposed a peak-piloted deep network (PPDN) for intensity-invariant expression recognition. %Specifically, the PPDN takes a pair of peak and non-peak expression images of the same type and from the same subject as input and utilizes the L2-norm loss to minimize the distance between both images. 
Moreover, Yu \textit{et al.}~\cite{yu2018deeper} proposed a deeper cascaded peak-piloted network (DCPN) that enhances the discriminative ability of features in a cascade fine-tuning manner. The PPDN and DCPN respectively achieved the best accuracies of 99.3\% and 99.9\% on the CK+ dataset~\cite{lucey2010extended}. However, these methods depend only on static frames and lack powerful modeling of the spatial-temporal relationships of expressions in the video. 

\begin{figure*}[htpb]
\begin{center}
	\includegraphics[width=1.0\linewidth]{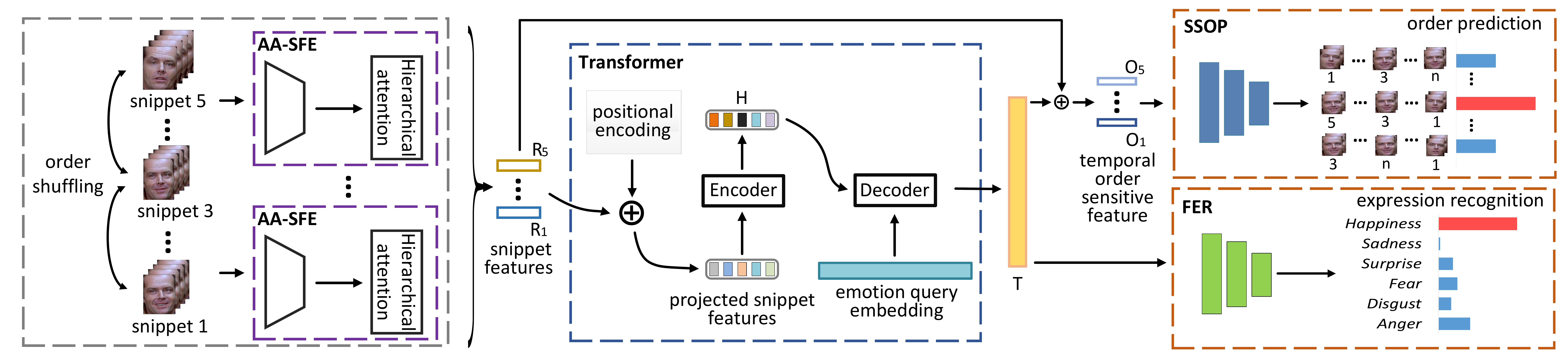}
\end{center}
\vspace{-0.4cm}
	\caption{The training pipeline of the EST for video-based FER. Using expression snippets, we apply the AA-SFE and SSOP head to improve Transformer's ability to model intra-/inter-snippet expression movements and relevance, thus achieving robust FER. }
\label{fig:TR-FER}
\vspace{-0.3cm}
\end{figure*}

\textbf{Dynamic sequence-based methods} In order to explore the spatial-temporal representation of expressions, dynamic sequence-based methods take a video sequence as a single input and utilize both textural information and temporal dependencies in the sequence for more robust expression recognition~\cite{jung2015joint,kim2017multi,kim2017deep,vielzeuf2017temporal,li2020deep}. Recently, the Long Short-Term Memory (LSTM) and C3D are two widely-used spatial-temporal methods. Vielzeuf \textit{et al.}~\cite{vielzeuf2017temporal} firstly used pre-trained VGG-Face to extract spatial features, then utilized LSTM layers to encode temporal dependencies in a sequence. Kim \textit{et al.}~\cite{kim2017multi} proposed a new spatio-temporal feature representation learning for FER by integrating C3D and LSTM networks, which is robust to expression intensity variations.  
Although the C3D networks can capture the spatial-temporal change of an expression, the C3D networks introduce expensive space- and computational complexity to learn subtle expression movements more effectively. 
\textbf{Transformer} Transformer was introduced by Vaswani \textit{et al.}~\cite{ISI:000452649406008} as a new attention-based building block for machine translation. Transformer included self-attention layers to scan through each token in a sequence and learn the tokens' relationships by aggregating information from the whole sequence, replacing RNNs in many tasks, such as natural language processing (NLP), speech processing, and computer vision~\cite{devlin2018bert,luscher2019rwth,synnaeve2019end,radford2019language,parmar2018image,dai2020up-detr}.
Recently,  Nicolas \textit{et al.} expanded the basic Transformer architecture to the field of object detection and proposed the DETR algorithm~\cite{carion2020end}. %DETR
Girdhar \textit{et al.} proposed an action Transformer to aggregate features from the spatial-temporal contexts around persons
for action recognition in a video~\cite{neimark2021video}. %Video Action Transformer Network
Transformer has been successfully applied for computer vision tasks, such as objection detection and action recognition. However, applying the vanilla Transformer to capture subtle expression movements in an untrimmed video is still challenging due to the noises and the limited motion variations within input frames. 
%We thus advance the Transformer by introducing a novel snippet feature extractor and a novel learning scheme to overcome these issues.

%used the basic Transformer to model the subtle long-term dependence among expression frames in a untrimmed video, which not only temporal relation but also spatial relation. To overcome this issue, we combine transformers and temporal attention-augmented embeddings to better model elusive facial expression cues. To the best of our knowledge, it is the first time that the expression snippet Transformer is proposed for video-based FER.

\section{Expression Snippet Transformer}
\label{sec:TR-FER}

%In this section, we first describe the pipeline of the proposed EST in subsection 3.1, and then discuss the EST architecture choices and loss functions in the following subsections 3.2 $\sim $ 3.4.
\subsection{EST architecture}
%{\color{red}
The overall EST architecture is illustrated in Fig.~\ref{fig:TR-FER}. Firstly, we collect expression snippets from the input video. For each snippet, we apply the AA-SFE to extract per-snippet features. Then, we employ a Transformer with a SSOP head to achieve robust expression understanding. In the following sections, we will subsequently explain the Expression snippets, Transformer, AA-SFE, and SSOP. 
%In the following sections, we will first discuss the Transformer architecture, the AA-SFE, and the SSOP, subsequently. 
%It contain four main stages: (1) decomposing an input FER video $\mathcal{C}$ into a series of smaller sub-videos as expression snippets $\mathcal{C}=\{C_1,C_2,......C_n\}$, where $C_i$ represents the $i$-th sub-video and $n$ is the total number of sub-videos; (2) a novel feature extractor (AA-SFE) to extract intra-snippet expression features 
%Besides this inference pipeline, we additionally devise an shuffled snippet order prediction (SSOP) head and its loss to help improve the modeling of inter-snippet visual relations with Transformer during training.
%}
%Fig.~\ref{fig:TR-FER} shows the overview of the proposed EST training pipeline.

\noindent\textbf{Expression snippets} We decompose the input video into a series of snippets to augment the Transformer's ability to model subtle visual changes within each snippet and across different snippets, respectively. 
Formally, given an input FER video $\mathcal{C}$, we decompose it into a series of smaller sub-videos: $\mathcal{C}=\{C_1,C_2,......C_n\}$, where $C_i$ represents the $i$-th sub-video and $n$ is the total number of sub-videos. Each sub-video $C_i$ refers to an expression snippet that contains several adjacent frames of the video. 
%Different snippets do not overlap with each other. 
All the snippets have the same length, and they follow consecutive orders along time. % along the time axis. 
%To enhance the Transformer's ability for modeling subtle visual changes across frames, we introduce the AA-SFE to help encode temporal-spatial changes within each snippet. 

%The features $\mathcal{R}$ are then sent for relation modeling. 
\noindent\textbf{Transformer Architecture} 
%Using the snippet features $\mathcal{R}$, we then apply a snippet-based Transformer to improve the modeling of inter-snippet visual motion changes for discovering a unified salient emotion representation $T$ for FER. %Different from the original study \cite{ISI:000452649406008}, we introduce two prediction heads with the Transformer, i.e., the SSOP head and FER head, %attempts to shuffle the order of snippets and train the Transformer to predict the shuffled order $O$ correctly, 
%which help Transformer obtain much more facial movement patterns in a video and accurate expression recognition.} %stacked encoders followed by stacked decoders to translate several snippet features into a unified salient emotion representation $T$. 
%Afterwards, the FER prediction head uses an additional multi-layer perception network to perform robust FER. } 
%\subsection{Transformer} 
%In the following sections, we will describe the feature extractor, the Transformer architecture, and the order shuffle and reconstruction mechanism in details.
%\subsection{Snippet-based Transformer}
%As shown in Fig.~\ref{fig:TR-FER}, using snippet features $\{R_i\}$, we apply an encoder-decoder Transformer to obtain robust and salient expression representation by modeling relations and inter-snippet visual changes. %The final output of the Transformer is called the unified salient emotion representation $T$.
%\textbf{Encoder}
We first extract snippet features with AA-SFE, which will be discussed later. 
With the snippet features, a Transformer is applied here to model the expression movements across snippets and discover a more robust emotion representation for FER. 
We follow the typical Transformer formulation and apply a multi-head attention-based encoder-decoder pipeline for the processing. In general, the multi-head attention estimates the correlation between a \textit{query} tensor and a \textit{key} tensor and then aggregates a \textit{value} tensor according to correlation results to obtain an attended output. For more details of the Transformer, please refer to \cite{ISI:000452649406008}. 

In our approach, we employ the encoder to encode snippet features and then use the decoder to translate the encoded features into a more robust expression representation. 
%Firstly, for each snippet, we apply AA-SFE to help extracting snippet feature.
Let $R_i\in \mathbb{R}^{d}$ denote the extracted snippet feature of $C_i$ with a size of $d$, and $\mathcal{R}= \{R_1,R_2,......R_n\}$.
%$n$ is the number of all snippets,
%Suppose that the extracted feature for $C_i$ is $R_i$, we then have the collection of snippet features $\mathcal{R}= \{R_1,R_2,......R_n\}$. Each snippet feature $R_i$ is a $d$-dimensional feature vector, 
We feed $\mathcal{R}$ to the encoder of the Transformer in EST. 
In the encoder, for each head of the multi-head attention, we perform linear projections on a snippet feature $R_i$ to obtain the corresponding query vector $q_i$, key vector $k_i$, and value vector $v_i$, respectively. All the $q_i,k_i,v_i$ are vectors of size $d$ as well.

%4first compute the query vector, the key vector, and the value vector for each snippet feature $R_i$ via fully-connected layers, respectively. %Fully-connected layers are employed to transform each $R_i\in \mathcal{R}$ into the three vectors, respectively.
%Note that we make all the snippet features in $\mathcal{R}$ share the same projection parameters. 
Then, we stack different snippets' query vectors, key vectors, and value vectors to obtain a query tensor $Q$, a key tensor $K$, and a value tensor $V$, respectively. $Q, K, V\in\mathbb{R}^{n\times d}$.
Afterward, we perform self-attention across the snippets based on the obtained $Q$, $K$, and $V$. In addition, we apply a snippet positional encoding to describe the positions of snippets within a video, following the sine and cosine positional encoding \cite{ISI:000452649406008}.  %order of snippets and 
%Following the Transformer architecture, the relations between any pair of snippets is modeled using self-attention. 
The output of the encoder is the encoded snippet features $H\in \mathbb{R}^{n \times d}$:
\begin{equation}
   H  = A({Q}, {K}, {V}) = softmax(\frac{{Q}  ~ {K}^T}{\sqrt{d}}){V},
    \label{eq:sa} 
\end{equation}
where $A(\cdot)$ represents the self-attention.
%Like the original study \cite{ISI:000452649406008}, we also apply multi-head attention. Formally, for $p$-th head output embedding $H_p$, it is computed via:
%\begin{equation}
%	H_p = A(Q, K, V)= softmax(\frac{{Q}~{K}^T}{\sqrt{d}}){V},
%	\label{eq:sa}
%\end{equation}
%where $A(\cdot)$ is the self-attention operation. $d$ is the dimension of the features $R$.
%where $A(\cdot)$ represents the self-attention operation.
%Then, all the $H_p$ are concatenated together to form the encoder final output.
In this study, we employ 3 encoder layers, each with 4 attention heads. 
%We also connect each encoder output to its input feature via two-layer residual structure and apply a normalization operation to regularize the features.  
%, and $A$ represents self-attention function:
%\begin{equation}
%A(Q, K, V)= softmax(\frac{Q \cdot K ^T}{\sqrt{d}})V,
%\end{equation}
%and $d$ is the dimension of both queries and keys. 
%\textbf{Transformer Decoder} 

After encoding snippet features with self-attention, the decoder phase then applies cross-attention to decode the encoded features $H$ into an emotion representation $T$ and $T\in\mathbb{R}^{d}$. %We define the unified salient emotion representation $T$ as the most important emotion cues discovered for expression recognition. 
We introduce an \emph{emotion query embedding} for the decoder to represent the query tensor of the multi-head attention. The emotion query embedding shares the same dimension with $T$. 
We use the encoded feature $H$ to represent both key and value tensors in the decoder. 
%一会儿是tensor一会儿是matrix，需要统一！
%For example, when we use 7 snippets and each has a 512-dimensional feature, the encoder also outputs a 7 x 512 feature map. Then, the decoder phase will search for a single 512-dimensional (1 x 512) feature vector that can better represent the input video using self-attention operations. To enable the effective decoding, we further introduce  for each employed decoder layer. The emotion saliency query has the same dimension with the output vector (1 x 512 using the above setting). %It can serve as a general description of what the salient emotion representation should be, 
%{\color{red}The salient emotion representation can be found according to the relations of snippet features to the emotion saliency query. }
%The relation of snippet feature is modeled as a 1x7-dimensional vector in the decoder that can describe the salient weight of each snippet. 
In practice, we stack 3 decoder layers, each with 4 attention heads, to progressively refine the decoding results. 

After the encoder-decoder processing, we make the Transformer provide two outputs, forming two prediction heads. The first head, built upon a 3-layer perception network, is the expression recognition prediction, classifying the $T$ into different expression types. The second head is the SSOP, which estimates the correct snippet order since snippets are shuffled. We will discuss the details of SSOP later. 
%We introduce the SSOP to help improve the modelling of inter-snippet relations with the Transformer. We will discuss its details later. 

%the Transformer contains two prediction heads to help improve the modelling of inter-snippet relations with the Transformer. %We predict the final facial expression of a video based on the output of the decoding phase. 
%The one is the expression prediction head, which is made upon a 3-layer perceptron network with ReLU activation function based on the output of the decoding phase. 
%The other is the SSOP head, which predicts inter-snippet movement changes via a self-supervision learning scheme.

\subsection{Attention-augmented Snippet Feature Extraction}

\begin{figure}[htp]
\begin{center}
	\includegraphics[width=1.0\linewidth]{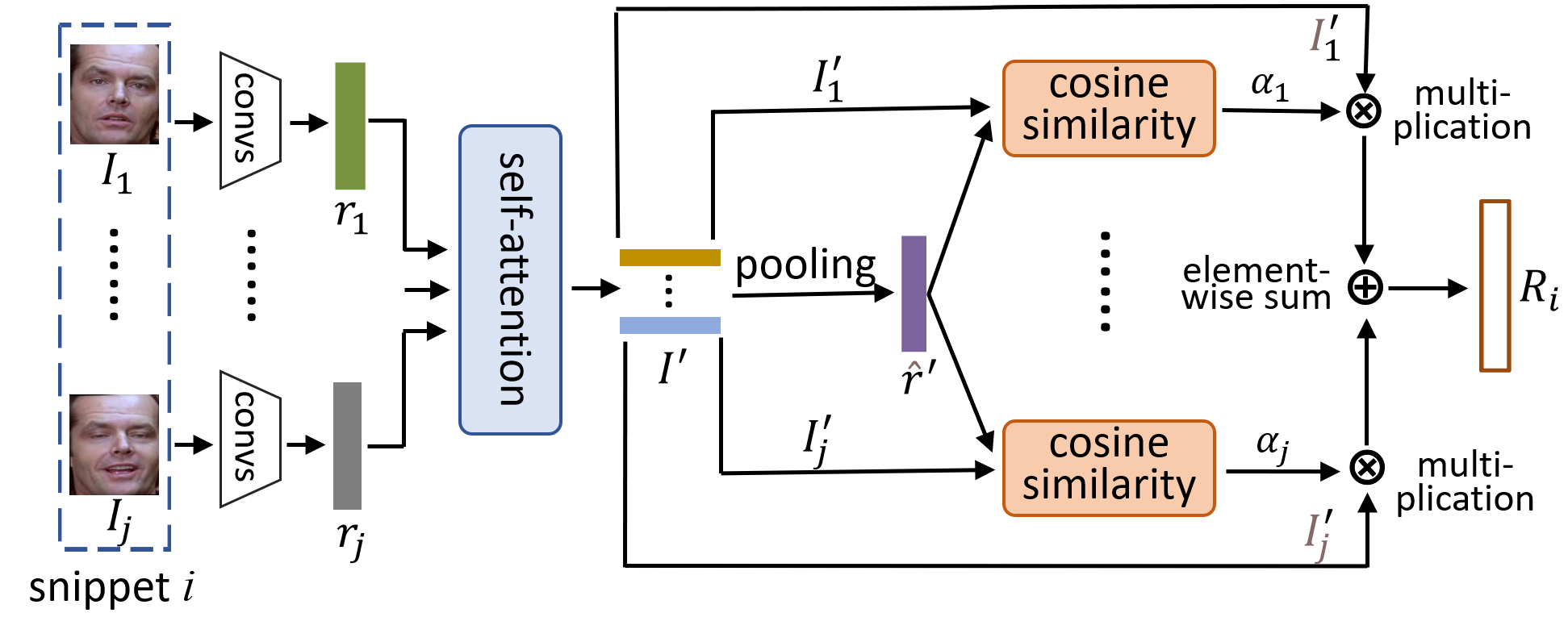}
\end{center}
	\vspace{-0.5cm}
	\caption{The detailed architecture of the AA-SFE. }
	\vspace{-0.2cm}
\label{fig:CFE}
\end{figure}

%Using the Transformer, we find that it is very difficult to effectively model relations between frames due to the extremely small intra-snippet visual changes, as well as the significant noises existed within raw pixels. By decomposing the video into snippets, we attempt to boost the modeling power of the Transformer to obtain a more robust representation of spatial-temporal information across frames in a snippet. 
Directly applying the Transformer on raw frames can be sub-optimal due to visual noises within pixels, making it difficult to obtain robust expression representation.
Using snippets, we boost the Transformer to better model intra-snippet expression movements by introducing a novel attention-augmented snippet feature extractor (AA-SFE). AA-SFE improves the encoding of spatial-temporal information across frames within a snippet. 
%In order to obtain snippet-based expression features from a untrimmed video, the AA-SFE component learns inter-frame and intra-frame attentions within each snippet via jointly frame attention learning (FAL) and frame relation learning (FRL). 

%\textbf{Hierarchical Attention-Augmentation} 
Fig.~\ref{fig:CFE} shows the structure of an AA-SFE. In particular, with the help of normal DCNNs, such as a pre-trained ResNet-18, the AA-SFE applies a \textit{hierarchical attention augmentation} for modeling intra-snippet information.
The hierarchical attention aims to gradually extract a more representative feature of a snippet, progressively filtering out less meaningful non-expression information to reduce the negative impacts of noises within per-frame features. We mainly apply the attention from \textit{two-level} hierarchy to model subtle visual changes. The first level extracts frame-level attention, and the second focuses on extracting snippet-level global attention. 

For the \textit{first-level} hierarchy, we investigate frame-level relation to obtaining attention. Similar to the Transformer, we apply self-attention here for relation modeling. 
Mathematically, we use $r_{j,i}$ to represent the feature vector of the $j$-th frame in the $i$-th snippet. We extract the global average pooling output of a DCNN as the per-frame feature: $r_{j,i} \in \mathbb{R}^d$. 
%We  and each $r_{j,i}$ has $d$ dimensions. 
Suppose each snippet has $J$ frames. 
By stacking all the features $r_{j,i}$ from the $i$-th snippet, we obtain the tensor $I_i\in\mathbb{R}^{J\times d}$.
Since we only consider frames of a single snippet at this stage, we drop the symbol $i$ here for simplicity, \textit{i.e.}, $I=I_i$, $r_j=r_{j,i}$ in this part.
Using linear projections, we transform $I$ into three tensors: query tensor $I_{Q}$, key tensor $I_{K}$, and value tensor $I_{V}$. Then, we apply self-attention described in Eq.\ref{eq:sa} on $I_Q, I_K, I_V$ to obtain the attended feature $I'\in\mathbb{R}^{J\times d}$.
%$  I'  = A(I_{Q}, I_{K}, I_{V})$,
% described as equation \ref{eq:sa}.}

%Meanwhile, we use $\alpha_{1,k}$ to denote the $k$-th frame-level self-attention, and it is computed according to the following operation:
%\begin{equation}
   % \alpha_{1,k}  = \frac{\rho_{1,k}}{\sum_k \rho_{1,k}}, ~~ \rho_{1,k} = \sigma(W_{1,k} I_i^f),
    %\alpha_{1,k} = softmax(\frac{Q_I \cdot K_I^T}{\sqrt{d}}),
    %\label{eq:a1-1}
%\end{equation}
%where $\alpha_{1,k}$ is the attentional score of per-frame feature and $d$ is the dimension of both queries and keys. Then, the per-frame features $I^f_{k,g}$ after modeling self-attention with other frames are calculated by:
%\begin{equation}
	%I^f_{k,g} = \sum_k \alpha_{1,k}\cdot I^f_k.
%\end{equation}

%Given a frame $l_k$ in a snippet $C_i$, FAL is used to explore the spatial characteristics of facial expressions via initially inter-frame attention learning. %Formally, let $C_k$ be the $k^{th}$ clip in a video,  which includes $j$ frames as:
%\begin{equation}
%	C_k=\{I_1,I_2,..., I_i,...,I_j\}
%\end{equation}
%where $ I_i$ is the $i^{th}$ frame in the clip.
%The frame-level feature $I_i^f$ is first extracted by a pre-trained ResNet~\cite{}, and then fed to a fully-connected layer (FC) and a sigmoid function to calculate inter-frame attentions. The inter-frame attention $\alpha_i$ is achieved by:
%\begin{equation}
%	\alpha_i = \sigma(I_i^f \cdot q^0)
%\end{equation}
%where $q^0$ is the parameter of the FC, $\sigma$ denotes the sigmoid function. 

%\textbf{Frame relation learning} 
In the \textit{second-level} hierarchy, we introduce the snippet-level global information to further refine representations of snippets. Firstly, we summarize the $I'$ into a unified general feature vector. Then, we estimate the relations between the general feature and per-frame features. The obtained relations are later used to re-weight per-frame features for refinement. 
Lastly, the refined features are reduced to a single representation to describe the whole snippet. 
More specifically, we denote the symbol $\hat{r}'$ as the general feature vector of $I'$, with a size of $d$. It is obtained by performing max pooling on $I'$ across frames. Then, we estimate the relation between $\hat{r}'$ and per-frame features using cosine similarity. 
%and then estimate its relation to each frame using cosine similarity. The general feature is a max pooling result of $I'$ performed across frames. 
%The estimated similarity scores serve as another attention to weight the importance of features from different frames. 
%Lastly, the weighted features are then reduced to a single representation to describe the whole snippet. 
%More specifically,  %The $\hat{r}'\in\mathbb{R}^{K\times d}$ is obtained by performing max-pooling on the per-frame features of $I'$, encoding global temporal representation for the snippet. 
We denote $r'_j$ as the feature in $I'$ correspond to the $j$-th frame. 
We compute cosine similarity $\alpha_j$ between $\hat{r}'$ and each $r'_j$:
\begin{equation}
    \alpha_j = cos(r'_j, \hat{r}') = \frac{r'_j \cdot  \hat{r}'}{\|r'_j\| \cdot \|\hat{r}'\|},
    \label{eq:cos} 
\end{equation}
where $\|\cdot\|$ means Euclidean norm. With the relation estimated by $\alpha_j$, we can identify which frame contains more deviated information that could be more likely to contain noise with non-expression. Thus, we have the summarized snippet feature by re-weighting and aggregating per-frame features based on:
\begin{equation}
    R_i = \frac{\sum_j \alpha_j \cdot r'_j}{\sum_j \alpha_j}.
    \label{eq:r}
\end{equation}

 %we first obtain an effective frame-level representation in a snippet by attending to other frames.} 
%{\color{red}
To sum up, the self-attention of AA-SFE first provides powerful relation modeling to facilitate the encoding of frame-level spatial-temporal information. %However, subtle inter-snippet visual changes and potential noises existed within each frame could still distract the self-attention.
Then, we introduce the second hierarchy with cosine similarity-based attention modeling to consider the global motion information of a snippet to help further resist noises existing in per frame. 
According to Eq.~\ref{eq:cos} $\sim$ \ref{eq:r}, the attention can identify the more useful intra-snippet visual change information and facilitate the computation of a more focused snippet feature $R_i\in\mathbb{R}^{d}$. %}
We experimentally prove that the AA-SFE delivers better snippet features comparing to a normal self-attention-based Transformer. 

\subsection{Shuffled Snippet Order Prediction }
%\subsubsection{Motivation}

%{\color{red}
%In practice, we observe that the facial expression movements across different snippets are still so small that the Transformer usually 
%More specifically, by investigating the attentional weight of each snippet, we find the Transformer easily learns to only focus on the snippet with peak expression changes,  
%we observe that the proposed EST easily overfits to the snippet order during training. 
%More specifically, by investigating the attentional weight of each snippet, we find that the Transformer easily learns to only focus on the information of the snippet at some fixed timestamps in a video, ignoring the actual visual changes. 
%We analyze that this is because informative emotion cues are always distributed around similar positions in a training video.

%With the snippet features $\mathcal{R}$ and the Transformer, we can estimate expressions of videos. %{\color{blue}However, we observe that the Transformer usually only focuses on the snippet with peak expression changes and neglects the rest parts of a video. We analyze that this due to its cross-attention modelling mechanism which mainly models relations of different snippets but can easily overlook the motion changes across subsequent snippets. In practice, the Transformer usually fails to deliver the comprehensive inter-snippet relation modeling for all the snippets and thus can be easily distracted by noisy information in the peak snippet.}
With the snippet features $\mathcal{R}$ and the Transformer, we can estimate expressions of videos. However, we observe that the Transformer usually focuses only on the snippet with peak expression changes and neglects the rest parts of a video. This situation happens because the cross-attention modeling mechanism of Transformer probably easily overlooks the slight motion changes across subsequent snippets. In practice, the Transformer usually fails to deliver the comprehensive inter-snippet relation modeling for all the snippets and thus can be easily distracted by noisy information in the peak snippet.
%which means that it usually abandons the inter-snippet relation modeling and can be disturbed by useless information in the peak snippet.
%} %is usually distributed around the medium of a FER video. 
%The rest parts of the video only contain very small facial changes, so that the Transformer usually learns to abandon the inter-snippet relation modeling. The Transformer would only focus on the middle snippet of $\mathcal{C}$ for recognition, ignoring plenty of beneficial visual information from other periods of an input video. We argue that this can greatly limit the capability of Transformer to recognize expressions robustly. 
To make the Transformer model expression motions more comprehensively and avoid the negligence of subtle visual changes from off-peak snippets, we further introduce a shuffled snippet order prediction (SSOP) head with corresponding loss to assist training the EST. 
The algorithm of the SSOP is shown in Algorithm \ref{alg:sosr}.

\begin{algorithm}[htb]
		% \algsetup{linenosize=\small}
		% \scriptsize
		\caption{The Pseudo Code of the SSOP.}
			\textbf{Input}:FER video $\mathcal{C};$Permuted order $S$\\
			\textbf{Output}:Predicted permuted order probability $p(S|O)$
			\begin{algorithmic}[1]
			%\Function{SOSR}{$\mathcal{C},S$}
			%\For{number of training iterations}
			%\For{$\mathbf{all}\ v_i \in V and\ l_j \in S$}
			\STATE $\{C_1,...,C_n\} \leftarrow \mathcal{C}$
			\STATE Shuffle the snippet order according to the $S$
			%\State $R \leftarrow AA-SFE(C)$
			%\State $D \leftarrow Transformer(R)$
			\STATE Collect featuers: $O_i = R_i + T$
			\STATE Concat features: $O = Concat([O_1,...,O_n])$ 
			\STATE Predict the snippet order probability $p(S|O)$
			\STATE Update SSOP head and emotion information $T$ by Eq.~\ref{eq:lrec} %descending stochastic gradient:\\
			%	$ \nabla_\theta[-\sum_{l \in S}{ l\cdot log(p(l_j|O))}]$
			%\EndFor
			%\EndFor
			%\EndFunction
			\end{algorithmic}
		\label{alg:sosr}
\end{algorithm}
%\vspace{-0.2cm}
% \end{tiny}
%\vspace{-0.2cm}

To train the Transformer with SSOP, we mainly shuffle the snippets and make the Transformer predict this shuffled snippet order. For example, if we have 7 snippets, a shuffled order can be like $S = (3, 2, 7, 1, 4, 6, 5)$. In practice, we generate 10 different types of different shuffled orders. 
Then, among all the generated orders, we sample one order and re-arrange the snippets accordingly. The snippets with a shuffled order are later sent to the EST. After extracting the emotion information $T$, we fuse the $T$ with the $\mathcal{R}\in\mathbb{R}^{7 \times d}$ to obtain the features used for predicting shuffled orders. We name this type of feature as temporal order sensitive feature $O\in\mathbb{R}^{7 \times d}$. The computation of $O$ can be found in Algorithm \ref{alg:sosr}. We further apply three fully connected layers on the $O$ to define the SSOP head, predicting the current permuted shuffling order.
The prediction is obtained according to a classification output. Therefore, training the Transformer with the SSOP involves maximizing a posterior probability (MAP) estimate, where the related conditional probability density function is:
\begin{equation}
	p(S|C_1, C_2, ..., C_n)=p(S |O_1,...,O_n)\prod_{i=1}^{n}p(O_i | C_i),
\end{equation}
where $O_i$ is the feature vector in $O$ for the $i$-th snippet. $C_i$ represents $i$-th snippet. 

In practice, without SSOP, although we have positional encoding in the Transformer, the snippet order and motion information from off-peak snippets is usually not well encoded due to very subtle facial changes. Alternatively, training the Transformer to identify shuffled snippet orders with SSOP can help ensure that information from every snippet is properly attended. As a result, the Transformer could become more sensitive to inter-snippet visual changes and more comprehensive to describe expression changes of the entire video. Besides, the SSOP also enriches the number of expression change patterns for training without requiring additional manual annotation.

\subsection{Optimization Objectives}
 For training, the EST has two objectives. The first one is %to predict the facial expression of a video via optimizing 
 a FER classification loss $L_{cls}$, and the second one is a shuffled snippet order prediction loss $L_{S}$. We use the cross-entropy loss for optimization.
 %reconstruct the snippets' temporal order and further model the most important emotion cues via optimizing a temporal reconstruction loss $L_{rec}$. %To achieve the above-mentioned two learning objectives during training, two joint loss functions are employed by the proposed EST, \textit{i.e.}, a FER loss and a snippet order reconstruction loss.
 Mathematically, the FER loss 
$ L_{cls} $ can be written as:
\begin{equation}
%{L_{cls}} =\frac{1}{{{N_{\rm{B}}}}}\sum\limits_{j = 1}^{N_{\rm{B}}} {{{({{\bf{\hat y}}_e} - {\bf{y}}_e)}^2}},   
{L_{cls}}\!=\!-\sum_{\mathcal{C}}{ {Y}_\mathcal{C}  \cdot log[\hat{Y}_\mathcal{C} ]\!+\!(1-{Y}_\mathcal{C})\cdot log[1\!-\!\hat{Y}_\mathcal{C}]},
\label{eq:lcls}
\end{equation}
where ${Y}_\mathcal{C} $ denotes the facial expression label for each video, $\mathcal{C}$ indexes a training video, and $\hat{{Y}}_\mathcal{C}$
%\in \mathbb{R}^E$ 
denotes the probabilities of facial expressions predicted by the EST. 
%$p(\hat{\mathbf{y}}|T)$ is a probability distribution of prediction.

To identity shuffled snippet order, we introduce the loss function $L_{S}$ for SSOP based on:
\begin{equation}
	L_{S}=-\sum_{\mathcal{C}}{ {S}_\mathcal{C} \cdot log[\hat{S}_\mathcal{C}]+(1-{S}_\mathcal{C})\cdot log[1-\hat{S}_\mathcal{C}]},
		\label{eq:lrec}
\end{equation}
where $\hat{S}_\mathcal{C}$ denotes the permutation type of the shuffled order predicted by the EST, and ${S}_\mathcal{C}$ is the ground truth one-vs-all label indicating the correct permutation type.  %order of the video. 
%$l_j$ denotes the permuted order prediction. The SOSR branch maps reconstructed the snippet order into a probability distribution $p(l_j|O)$.

\section{Experimental Results}\label{sec:Exp}

\subsection{Datasets}
To evaluate our approach, four face expression datasets were used: BU-3DFE dataset~\cite{yin20063d}, MMI dataset~\cite{valstar2010induced},  AFEW8.0 dataset~\cite{dhall2015video}, and DFEW dataset~\cite{jiang2020dfew}. 
%\textbf{BU-3DFE~\cite{yin20063d}:} The 3D facial expressions are captured at a video rate (25 frames per second). Six emotion labels are included, \textit{i.e.}, anger, disgust, happiness, fear, sadness, and surprise. Each expression sequence contains about 100 frames. BU-3DFE contains 606 3D facial expression sequences captured from 101 subjects, with a total of approximately 60,600 frames. %We conducted a 10-fold validation on the dataset.%,  where 486 videos are used for training and 120 videos for testing.%Each 3D model of a 3D video sequence has the resolution of approximately 35,000 vertices. The texture video has a resolution of about 1040$\times$329 pixels per frame. The resulting database consists of 58 female and 43 male subjects, with a variety of ethnic racial ancestries, including Asian, Black, Hispanic, Latino, and White.

\textbf{BU-3DFE~\cite{yin20063d}:} 3D facial expressions annotated with 6 emotion labels, \textit{i.e.}, anger, disgust, happiness, fear, sadness, and surprise. BU-3DFE contains 606 3D facial expression sequences captured from 101 subjects. Each expression sequence contains nearly 100 frames. 

\textbf{MMI~\cite{valstar2010induced}:} A total of 205 expression sequences were collected from 30 subjects. The expression sequences were recorded at a temporal resolution of 24 fps. Each expression sequence of the dataset was labeled with one of the six basic expression classes (\textit{i.e.}, anger, disgust, fear, happiness, sadness, and surprise). %The expression sequences were collected such that, the first frame in the sequence was the onset frame and last frame was the offset frame. %The indexes of the apex frames were located manually. 
 %Additionally, a 10-fold validation was conducted on the dataset. %171 videos and 37 videos were respectively used for training and testing.

\textbf{AFEW~\cite{dhall2015video}:} The AFEW serves as an evaluation platform for the annual EmotiW since 2013. Seven emotion labels are included in AFEW, \textit{i.e.} anger, disgust, fear, happiness, sadness, surprise, and neutral. AFEW contains videos collected from different movies and TV serials with spontaneous expressions, various head poses, occlusions, and illuminations. AFEW is divided into 
three splits: Train (738 videos), Val (352 videos), and Test (653 videos). %We do not have test labels for evaluation. %Hence, we follow the setting of other compared methods and only used the Train/Val set for experiments.%In this study, we only used the Training and Val set.

\textbf{DFEW~\cite{jiang2020dfew}:}  The DFEW is a large-scale unconstrained dynamic facial expression database, containing 16,372 video clips extracted from over 1,500 different movies. It contains 12,059 single-label video clips and also includes seven emotion labels, \textit{i.e.} anger, disgust, fear, happiness, sadness, surprise, and neutral. 
%DFEW dataset provides a 5-fold validation for evaluation. %includes a training set (9,356 videos) and test set (2,341 videos), and 

\subsection{Snippet Extraction and Implementation Details}
We first unified the video length to 105 frames via interpolation and clipping operation, and detected face regions of each frame to the size of 224$\times$224 via the Retinaface~\cite{deng2020retinaface}. %During the training phase,
Then, we randomly selected one of the first 30 frames as the starting frame, and extracted the following 75 consecutive frames to form a video. Next, we split the 75 frames into 7 sub-videos, each of which had 15 frames, with five frames overlapping between each sub-video. To enhance expression movement variation, 5 frames were randomly sampled from each sub-video to form a new sub-video which is an expression snippet. Thus, $n=7$ and $j=5$.
For training, the seven snippets were shuffled in a random order (the frame order within each snippet remained unchanged). For test, %we first start the frame 15 and select 75 frames. Second, we split the 75 frames into seven snippets, each of which has 15 frames, with five frames overlapping between each snippet. Third, we select one frame in every three frames of each snippet as test data. Finally, 
 we only used the normal snippet order as input for robust FER. 
%the size of the temporal window is set as 50, and the step size is 10. For one temporal window, the 50 frames were divided into 5 groups on average, and three frames were randomly selected from the first, third and fifth groups. Therefore, each video can be divided into 7 clips and each clip contains 3 frames. The dimension of each frame is resized to 224$\times$224.

We used the Pytorch for implementing the EST. The key training parameters include initial learning rate (0.0001), cosine annealing schedule to adjust the learning rate, mini-batch size (8), and warm up. The experiments were conducted on a PC with Intel(R) Xeon(R) Gold 6240C CPU at 2.60GHz and 128GB memory, and NVIDIA GeForce RTX 3090. Following the setting of other compared methods, we conducted a 10-fold person-independent validation on the BU-3DFE and MMI, a Train/Val set validation on the AFEW, and a 5-fold validation on DFEW dataset. We will release our source code to Github after acceptance.

\subsection{Experiments on the BU-3DFE Dataset}
\begin{figure*}[htpb]
\begin{center}
	\includegraphics[width=1\linewidth]{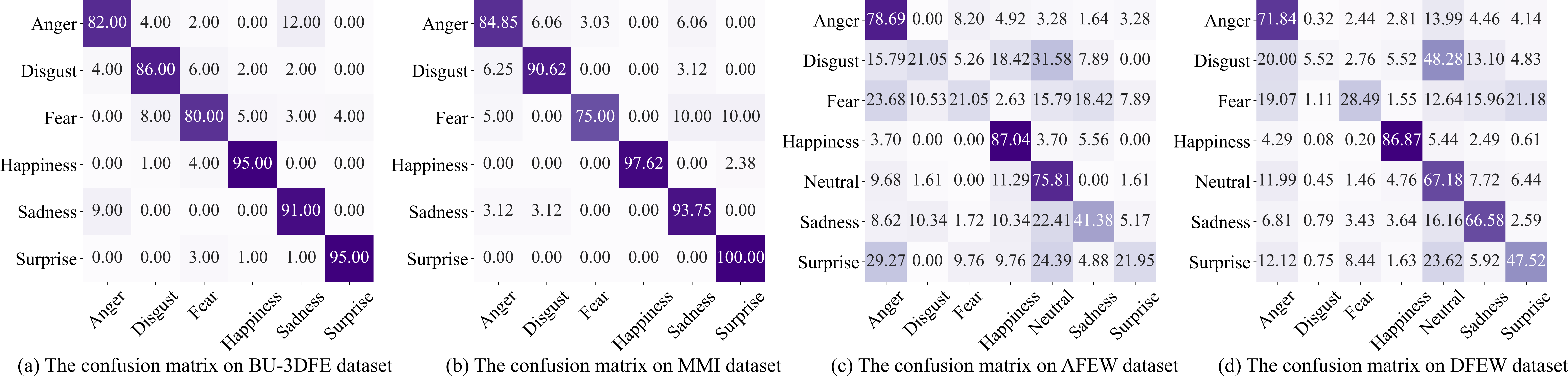}
\end{center}
\vspace{-0.4cm}
	\caption{The confusion matrixes for video-based FER on the four datasets.}
\label{fig:confusion_matrix}
\vspace{-0.3cm}
\end{figure*}

% \begin{figure}
% \begin{center}
% 	\includegraphics[width=0.7\linewidth]{fig/BU3D_confusion2.png}
% \end{center}
% 	\caption{The confusion matrix for video-based FER on the BU-3DFE dataset.}
% \label{fig:BU3D_confusion}
% \end{figure}

Fig.~\ref{fig:confusion_matrix}(a) shows the confusion matrix of BU-3DFE for video FER by using our method. Among the six expressions, the highest accuracy are 95.0\% of Happiness and Surprise, while the lowest accuracy is 80.0\% for Fear, which has the least amount of facial expression movement and is difficult to distinguish with Disgust. The average FER accuracy is 88.17\%.

The average FER accuracy of the EST was compared with the state-of-the-art methods, including DeRL~\cite{yang2018facial}, FAN~\cite{meng2019frame}, ICNP~\cite{2016Muscular}, C3D~\cite{2015Learning}, 2D+3D model~\cite{ly2019novel}, FERAtt+Rep+Cls~\cite{marrero2019feratt} and C3D-LSTM~\cite{parmar2017learning} in Table~\ref{tab:BU3D_compare}. Compared to the best sequence-based result, the proposed EST improved the accuracy over 4.97\%. This reveals that our method can effectively discover the more beneficial emotion-related cues by  modeling the long-range emotion movement relation in videos.

\begin{table}[htpb]
	\centering
	\resizebox{1.0\linewidth}{!}
	{
    	\begin{tabular}{l|l|c}
    		\hline
    		Methods & Feature setting & Accuracy(\%)
    		\\
    		\hline\hline
    		% MPCNN~\cite{liu2018multi} & peak frame-based & 91.22\\
    		FAN~\cite{meng2019frame} & frame-based & 84.17\\
    		2D+3D model~\cite{ly2019novel} & frame-based & 87.66 \\
    		FERAtt+Rep+Cls~\cite{marrero2019feratt} & frame-based & 82.11 \\
    		DeRL~\cite{yang2018facial} & peak frame-based & 84.17\\
    % 		CNN~\cite{yang2018facial} & peak frame-based & 73.20\\
    		\hline
    		C3D~\cite{2015Learning} & sequence-based & 82.18\\
    		ICNP~\cite{2016Muscular} & sequence-based & 83.20\\
    		C3D-LSTM~\cite{parmar2017learning} & sequence-based & 79.17\\
    		\hline
    		\textbf{Our EST} & \textbf{snippet-based} & \textbf{88.17}\\
    		\hline
    	\end{tabular}
	}
\vspace{-0.3cm}
	\caption{Comparison results on the BU-3DFE dataset. Note: the best result is highlighted in bold.}
	\label{tab:BU3D_compare}
\end{table}
\vspace{-0.3cm}

\subsection{Experiments on the MMI Dataset}
% \begin{figure}
% \begin{center}
% 	\includegraphics[width=0.7\linewidth]{fig/MMI_confunsion.png}
% \end{center}
% 	\caption{The confusion matrix for video-based FER on MMI dataset.}
% \label{fig:MMI_confusion}
% \end{figure}

Fig.~\ref{fig:confusion_matrix}(b) depicts the confusion matrix of  MMI for video FER by using our method. We achieved 100\% accuracy in Surprise category. The average accuracy of FER is 92.5\%.

In comparison with the state-of-the-art video-based FER methods, Table~\ref{tab:MMI_compare} lists the average accuracy on the MMI dataset using deep learning-based methods with spatial feature representation (\textit{i.e.}, AUDN~\cite{liu2015inspired}, DeRL~\cite{yang2018facial}, LSTM~\cite{kim2017multi}, Deep generative-contrastive networks (DGCN)~\cite{kim2017deep}, Ensemble Network~\cite{sun2019deep}, SAANet~\cite{liu2020saanet}, WMCNN-LSTM~\cite{zhang2020facial}, WMDCNN~\cite{zhang2020facial}), hand-crafted feature based methods (\textit{i.e.}, collaborative expression representation (CER) extracted from the apex frames and LPQ-TOP~\cite{jiang2013dynamic} extracted from the whole sequence), and our EST. As shown in the table, the proposed EST outperformed existing state-of-the-art FER methods. Compared to the second best method, Ensemble Network~\cite{sun2019deep}, the EST improved the accuracy of 1.04\%. %Our EST achieved the best performance rather than requiring any complex network integration.
%However, it demands lots of computing resources for network ensemble. 

\begin{table}[htb]
	\centering
	\small
	\resizebox{1.0\linewidth}{!}
	{
    	\begin{tabular}{l|l|c}
    		\hline
    		Methods & Feature setting & Accuracy(\%)
    		\\
    		\hline\hline
    		DeRL~\cite{yang2018facial} & frame-based & 73.23\\
    		WMDCNN~\cite{zhang2020facial} & frame-based & 78.20 \\
    		AUDN~\cite{liu2015inspired} & peak frame-based & 75.85\\
    % 		LBP+SRC~\cite{huang2010new} & peak frame-based	& 59.18\\
    % 		LPQ+SRC~\cite{wang2012facial} & peak frame-based & 62.72\\
    		CER~\cite{lee2016collaborative}  & peak frame-based & 70.12\\
    		Ensemble Network~\cite{sun2019deep} & peak+neutral frame & 91.46\\
    		\hline
    		LSTM~\cite{kim2017multi} & sequence-based & 78.61\\
    		DGCN~\cite{kim2017deep} & sequence-based & 81.53\\
    		LPQ-TOP+SRC~\cite{jiang2013dynamic} & sequence-based & 64.11\\
    % 		LBP-TOP+SRC~\cite{zhao2007dynamic} & sequence-based & 61.19\\
    		SAANet~\cite{liu2020saanet} & sequence-based & 87.06\\
    		WMCNN-LSTM~\cite{zhang2020facial} & sequence-based & 87.10\\
    		\hline
    		\textbf{Our EST} & \textbf{snippet-based} & \textbf{92.50}\\
    	    \hline
    	\end{tabular}
	}
\vspace{-0.3cm}
	\caption{Comparison results on the MMI dataset. Note: the best result is highlighted in bold.}
	\label{tab:MMI_compare}
\vspace{-0.2cm}
\end{table}
\vspace{-0.3cm}

\subsection{Experiments on the AFEW Dataset}
% \begin{figure}
% \begin{center}
% 	\includegraphics[width=0.8\linewidth]{fig/AFEW_confusion2.png}
% \end{center}
% 	\caption{The confusion matrix for video-based FER on AFEW dataset.}
% \label{fig:AFEW_confusion}
% \end{figure}

 Fig.~\ref{fig:confusion_matrix}(c) shows the confusion matrix of FER on the challenging AFEW dataset. The average accuracy of FER achieved 54.26\%. The highest accuracy is 87.04\% of Happiness followed by Anger and Neutral, which respectively reach 78.69\% and 75.81\%. 
 %We also observed that our method achieved a similar phenomenon to the other two methods~\cite{liu2018multi-feature,vielzeuf2017temporal}, with lower accuracies on Disgust and Fear. But even so, our method still achieved a current best result of 21.05\% in the two categories.
 Although accuracies of Disgust and Fear are relatively lower than the other categories, our method still out-performs other methods in recognizing both emotions.
 This may be caused by better modeling the relations of subtle expression movements between snippets.
Table~\ref{tab:AFEW_compare} reports the accuracies using the EST and state-of-the-art methods. It demonstrates that our method achieves the best performance with great robustness, meanwhile, has obvious advantages over other algorithms on the in-the-wild expression dataset. 

\begin{table}[htpb]
	\small
	\centering
	\resizebox{1.0\linewidth}{!}
	{
    	\begin{tabular}{l|l|c}
    		\hline
    		Methods & Feature setting & Accuracy(\%) \\
    		\hline\hline
    		FAN~\cite{meng2019frame} & frame-based & 51.18\\
    		HoloNet~\cite{yao2016holonet} & frame-based & 44.57\\
    		DSN-HoloNet~\cite{hu2017learning} & frame-based & 46.47\\
    		DSN-VGGFace~\cite{fan2018video} & frame-based & 48.04\\
    		\hline
    		C3D~\cite{2015Learning} & sequence-based & 30.11\\
    		DenseNet-161~\cite{liu2018multi-feature} & sequence-based & 51.44\\
    		VGG16+TP+SA~\cite{aminbeidokhti2019emotion} & sequence-based & 49.00\\
    		Emotion-BEEU~\cite{kumar2020noisy} & sequence-based & 52.49\\
    		Mode variational LSTM~\cite{baddar2019mode} & sequence-based & 51.44\\
    		Former-DFER~\cite{zhao2021formerdfer} & sequence-based & 50.92\\
    		\hline
    		\textbf{Our EST} & \textbf{snippet-based} & \textbf{54.26}\\
    		\hline
    	\end{tabular}
	}
\vspace{-0.3cm}
	\caption{Comparison results on AFEW 8.0 dataset. Note: the highest result is highlighted in bold.}
	\label{tab:AFEW_compare}
\end{table}
\vspace{-0.3cm}

\subsection{Experiments on the DFEW Dataset}
% \begin{figure}
% \begin{center}
% 	\includegraphics[width=0.8\linewidth]{fig/DFEW_confusion.png}
% \end{center}
% 	\caption{The confusion matrix for video-based FER on DFEW dataset.}
% \label{fig:DFEW_confusion}
% \end{figure}
Fig.~\ref{fig:confusion_matrix}(d) shows the confusion matrix of FER on the large-scale DFEW dataset. The average accuracy of FER achieved 65.85\% by using our method. The highest accuracy is 86.87\% of Happiness followed by Anger, which achieves 71.84\%. Although we only achieved 5.52\% accuracy in the Disgust category due to the huge imbalance of categories in the DFEW (only occupies 1.22\% in the DFEW dataset) , the compared results in Table~\ref{tab:DFEW_compare} shows that our method is still far superior to other algorithms. %This may be caused by an extreme imbalance of categories, because the Disgust category only occupies 1.22\% in the DFEW data set.
More detailed comparison results can be shown in Table~\ref{tab:DFEW_compare}. Compared to the state-of-the-art methods reported in ~\cite{jiang2020dfew}, the FER accuracy of our EST achieved significant improvement (over 9.34\%). 

\begin{table}[thp]
	\small
	\centering
	\resizebox{1.0\linewidth}{!}
	{
    	\begin{tabular}{l|l|c}
    		\hline
    		Methods & Feature setting & Accuracy(\%)
    		\\
    		\hline\hline
    		3D ResNet-18,EC-STFL~\cite{jiang2020dfew} & sequence-based & 56.51\\
    		C3D,EC-STFL~\cite{jiang2020dfew} & sequence-based & 55.50\\
    		P3D,EC-STFL~\cite{jiang2020dfew} & sequence-based & 56.48\\
    		R3D18,EC-STFL~\cite{jiang2020dfew} & sequence-based & 56.19\\
    		VGG11+LSTM,EC-STFL~\cite{jiang2020dfew} & sequence-based & 56.25\\
    		\hline
    		\textbf{Our EST} & \textbf{snippet-based} & \textbf{65.85}\\
    		\hline
    	\end{tabular}
	}
\vspace{-0.3cm}
	\caption{Comparison results on DFEW dataset. 
	%Note: %we only compare to the best single models and the highest result is highlighted in bold.
	} 
	\label{tab:DFEW_compare}
\vspace{-0.3cm}
\end{table}

\subsection{Ablation experiment and analysis}
To better understand the role of each module in the proposed EST, %we evaluated and visualized each component for video-based FER as follows. 
Table~\ref{tab:ablation_aasfe_sosr} presents the ablation results of the gradual addition AA-SFE and SSOP  components to the baseline Transformer framework. The Transformer achieved a video-based FER accuracy of 85.60\% on the BU-3DFE dataset. The further integration of AA-SFE improved the accuracy to 87.12\%, as the AA-SFE aids in the extraction of snippet-level features via jointly hierarchical attentions. Thanks to learning the order sensitive representation, the addition of SSOP resulted in an increase of 1.05\%.
%\subsubsection{Effect of components in our framework}

\begin{table}[t!]
	\footnotesize
	\centering
		\resizebox{1.0\linewidth}{!}
	{
	\begin{tabular}{ccc|cc|c}
		\hline
		Transformer & AA-SFE & SSOP & Params(M) & MACs(G) & Acc(\%)
		\\
		\hline\hline
		\checkmark  & & & 34.37 & 63.85  &85.60\\
		\checkmark & \checkmark & & 34.37 & 63.88 & 87.12 \\
		\checkmark &  \checkmark  & \checkmark & 42.78 & 63.89 & \textbf{88.17}\\
		\hline
	\end{tabular}}
%	\resizebox{\linewidth}{!}
%	{
%    	\begin{tabular}{ccc|cc|c}
%        	\hline
%        	Transformer & SOSR & AA-SFE & Params(M) & GFlops & Acc(\%)
%        	\\
%        	\hline\hline
%        	\checkmark  & & & 1540.65 & 63.70  &85.83\\
%        	\checkmark & \checkmark & & 1572.7 & 63.71 & 86.67 \\
%        	\checkmark &  \checkmark  & \checkmark & 1572.72 & 63.71 & \textbf{90.83}\\
%        	\hline
%    	\end{tabular}
%	}
\vspace{-0.3cm}
\caption{Ablation study of the proposed EST. Impact of integrating our different components (AA-SFE and SSOP) into the baseline Transformer on the BU-3DFE dataset.}% The best result is in bold.}
	\label{tab:ablation_aasfe_sosr}
\end{table}

%\begin{figure}[htpb]
%\begin{center}
%	\includegraphics[width=1.0\linewidth]{fig/loss_acc.png}
%\end{center}
%	\caption{Ablation study of EST. The curve of training loss and test accuracy with different components.}
%\label{fig:loss_acc}
%\end{figure}

 %Additionally, Fig.~\ref{fig:loss_acc} shows the training and prediction procedure with the increasement of epochs. The gradual addition AA-SFE and SOSR improved the Transformer performance on both training speed and stability.% (a), the addition of SOSR converges faster than baseline Transformer. The further integration of AA-SFE obtains faster convergence rate. In Fig.~\ref{fig:loss_acc} (b), the addition of SOSR can achieve high accuracy faster than baseline Transformer on the Test set. After the integration of AA-SFE, our EST can get higher accuracy faster. Furthermore, 
Furthermore, 
%to further analyse the contribution of each attention in AA-SFE, 
Table~\ref{tab:ablation_aasfe_aatention} lists the recognition results with different attention selection in the AA-SFE. Obviously, two-level hierarchical attention used in AA-SFE achieved the best performance without any computational cost, helping to describe more informative snippet features. 
\begin{table}[t!]
	\footnotesize
	\centering
	\resizebox{1.0\linewidth}{!}
	{
    	\begin{tabular}{c|c|c}
    		\hline
    	    Different attention & Params(M) & Acc(\%)
    		\\
    		\hline\hline
    		%Frame attention~\cite{meng2019frame} & 1586.92 & 90\\
    		w/o attention  &   34.37 &  85.60\\
    		%softmax attention  & 42.78 & 86.61\\
    		self-attention & 42.78 & 87.63\\
    	    SE-like attention~\cite{hu2018squeeze} & 43.30 & 87.46\\
    		%self-attention & self-attention & 42.78 & 88.14\\
    	    Our hierarchical attention & 42.78 & \textbf{88.17}\\
    		\hline
    	\end{tabular}
	}
\vspace{-0.3cm}
		\caption{Ablation study of different attention selection in AA-SFE. The best results are highlighted in bold.}
	\label{tab:ablation_aasfe_aatention}
\vspace{-0.3cm}
\end{table}

%\subsubsection{Visualization results of emotion cue distribution}

\begin{figure}[t!]
\begin{center}
	\includegraphics[width=1.0\linewidth]{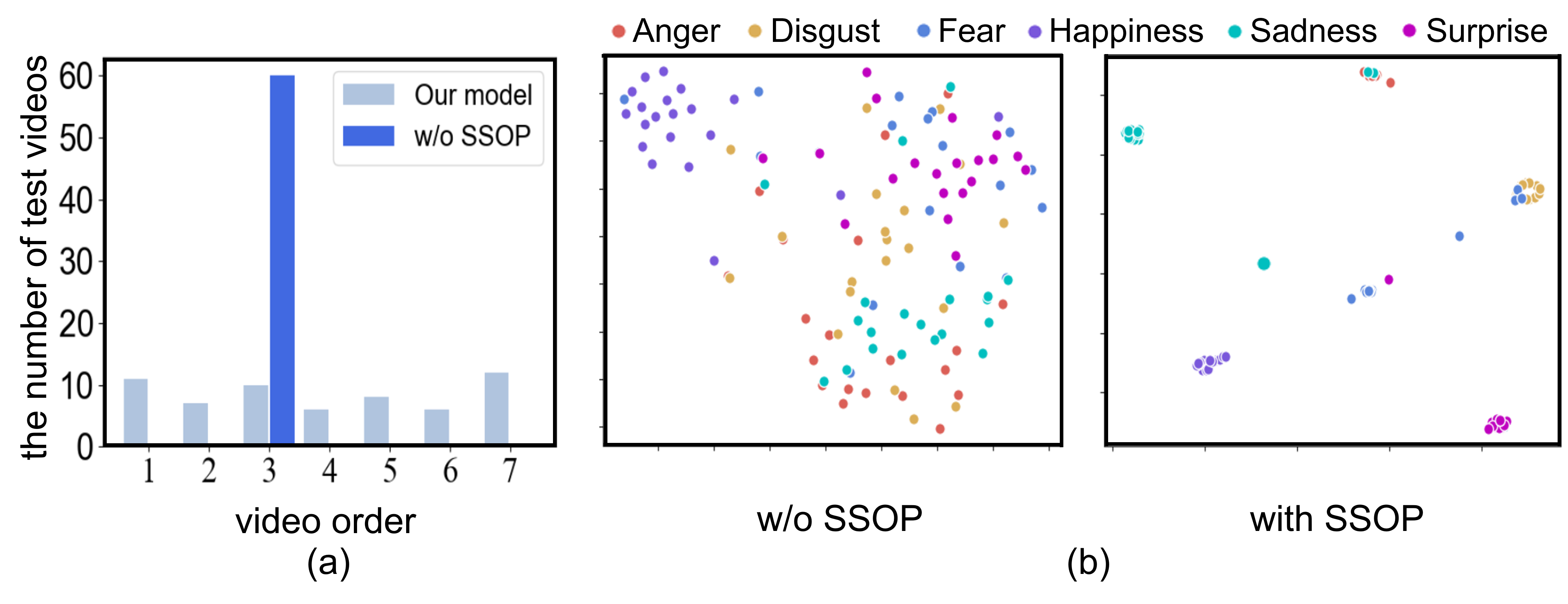}
\end{center}
\vspace{-0.5cm}
	\caption%Visualization in salient snippet location and salient weight learning whether using SOSR. 
	{The effects of SSOP.
	(a) Comparison of index distributions of the snippet with the highest attention weight with and without SSOP in EST; (b) Comparison of t-SNE analysis with and without SSOP. } %	most salient snippets identified by the EST with or without SSOP in shuffled order videos; 
	%(b) the emotion relation curve of snippets modeled in decoder of our model.} %It proved that the proposed SOSR branch can solve the overfit to order in Transformer and better describe the salient weight of each snippet in a video.}
\label{fig:visualization_result}
\vspace{-0.2cm}
\end{figure}

%\subsection{Effect of SSOP head}
In addition, Fig.~\ref{fig:visualization_result} shows more analysis about the effect of the SSOP in the EST. In particular, Fig.~\ref{fig:visualization_result}(a) presents the distribution of the index of the snippet with the highest attention weight in EST with and without SSOP, respectively. Without SSOP (see the dark-blue column in Fig.~\ref{fig:visualization_result}(a)), we can observe that the EST always focused on the 3-rd snippet, which usually contains the peak changes in each test video. Alternatively, the SSOP can make EST distribute similar attention to all the snippets. We further illustrate the encoded emotion representation in 2D space using t-SNE visualization with and without using SSOP in Fig.~\ref{fig:visualization_result}(b). The results show that the SSOP helps obtain more discriminative representation by comprehensively making the Transformer model inter-snippet visual changes.

\textbf{Model complexity}
Table~\ref{tab:comparison_complexity_efficiency} reports model parameters and computational costs of three spatial-temporal learning methods on the AFEW dataset. In general, our EST has the best performance (accuracy of 54.26\%) with a small computational cost (63.89G MACs) and real-time speed (412 fps), which means that the proposed method exhibits improved accuracy and efficiency. %Note that only using Transformer can achieve the accuracy of xxx due to lots of raw pixel noises and training over-fitting.
More ablation studies and discussions can be seen in the supplementary material.
 
\begin{table}[t!]
	\footnotesize
	\centering
		\resizebox{\linewidth}{0.8cm}{
	\begin{tabular}{l|l|l|ccc|c}
		\hline
		Methods &Input & Backbone & Params(M) & MACs(G) & fps & Acc 
		\\	
		\hline\hline
			C3D-LSTM~\cite{parmar2017learning} & Video &  ResNet-18 & 110.24 & 282.26  & 708 & 29.83 \\
		FERAtt~\cite{marrero2019feratt} & Frame  & ResNet-18 & 67.08 & 13.56 & 75   & 37.22 \\
	    Dense161~\cite{liu2018multi-feature} & Video & DenseNet-161 & 26.52 & 272.47 & 47 & 51.44\\
		VGG16TPSA~\cite{aminbeidokhti2019emotion} & Video & VGG16 & {14.72} & 537.61 & 552 & 49.00 \\
		\textbf{Our EST} & Video &{ResNet-18} & {42.78} & 63.89 & 412  & {\bf 54.26}\\
		\hline
	\end{tabular}
	}
\vspace{-0.3cm}
	\caption{Comparison of model complexity and efficiency.}
	\label{tab:comparison_complexity_efficiency}
\vspace{-0.3cm}
\end{table}

\section{Conclusions and Future Works}\label{sec:Conc}
In this paper, a novel expression snippet Transformer (EST) is proposed to better model elusive facial expression cues for robust facial expression recognition in untrimmed videos. The EST consists of four major components, \textit{i.e.}, snippet decomposition, snippet-based feature extractor, the encoder-decoder based Transformer, the shuffled order prediction head. Due to effectively and efficiently modeling long-range expression spatial-temporal relations and subtle intra-/inter- snippet visual changes, the proposed method achieved highly improved performance and strong robustness for video-based FER; the highest accuracies respectively reached 88.17\%, 92.5\%, 54.26\%, and 65.85\% on four challenging datasets (BU-3DFE, MMI, AFEW, and DFEW). In the future, we will introduce self-supervised learning to Transformer to model the extraction of emotion-rich features from complex unlabelled videos.

{\small
\bibliographystyle{plain}
\bibliography{ms}
}

\end{document}

\title{Supplementary Material for ``Expression Snippet Transformer for Robust Video-based Facial Expression Recognition"}
	\author[1]{Yuanyuan Liu}
	\author[1]{Wenbin Wang}
	\author[1]{Chuanxu Feng}
	\author[1]{Haoyu Zhang}
	\author[2]{Zhe Chen\footnote{Corresponding author}}
	\author[3]{Yibing Zhan}
	\affil[1]{China University of Geosciences (Wuhan)}
	\affil[1]{\textit{\{liuyy, wangwenbin, fcxfcx, zhanghaoyu\}@cug.edu.cn}}
	\affil[2]{The University of Sydney}
	\affil[2]{\textit{zhe.chen1@sydney.edu.au}}
	\affil[3]{Jingdong}
	\affil[3]{\textit{zhanyibing@jd.com}}

% REMOVE THIS: bibentry
% This is only needed to show inline citations in the guidelines document. You should not need it and can safely delete it.
\maketitle

%%%%%%%%% TITLE
%\title{Supplementary Material for ``Expression Snippet Transformer for Robust Video-based Facial Expression Recognition"}

%% omit the following lines up until the closing ``}''.
%% Additional authors and addresses can be added with ``\and'',
%% just like the second author.
%% To save space, use either the email address or home page, not both
%\and

%\maketitle
%\ificcvfinal\thispagestyle{empty}\fi

%%%%%%%%% ABSTRACT

%%%%%%%%% BODY TEXT
The content of our supplementary material is organized as follows.
\begin{enumerate}[1.]
	\item The training loss and testing accuracy of the EST with different components.
	\item Effect of important parameters in our framework.
	\item Compare with different backbones in our framework.
	\item Visualization of different video expression representations.
	\item Visualization of attention weights modelled by our EST and the Transformer.
	\item {Detailed description of the Transformer architecture used in our EST.}
\end{enumerate}

\section{Training Loss and Testing Accuracy Curves for different component study}

\begin{figure}[thpb]
	\begin{center}
		\includegraphics[width=1.0\linewidth]{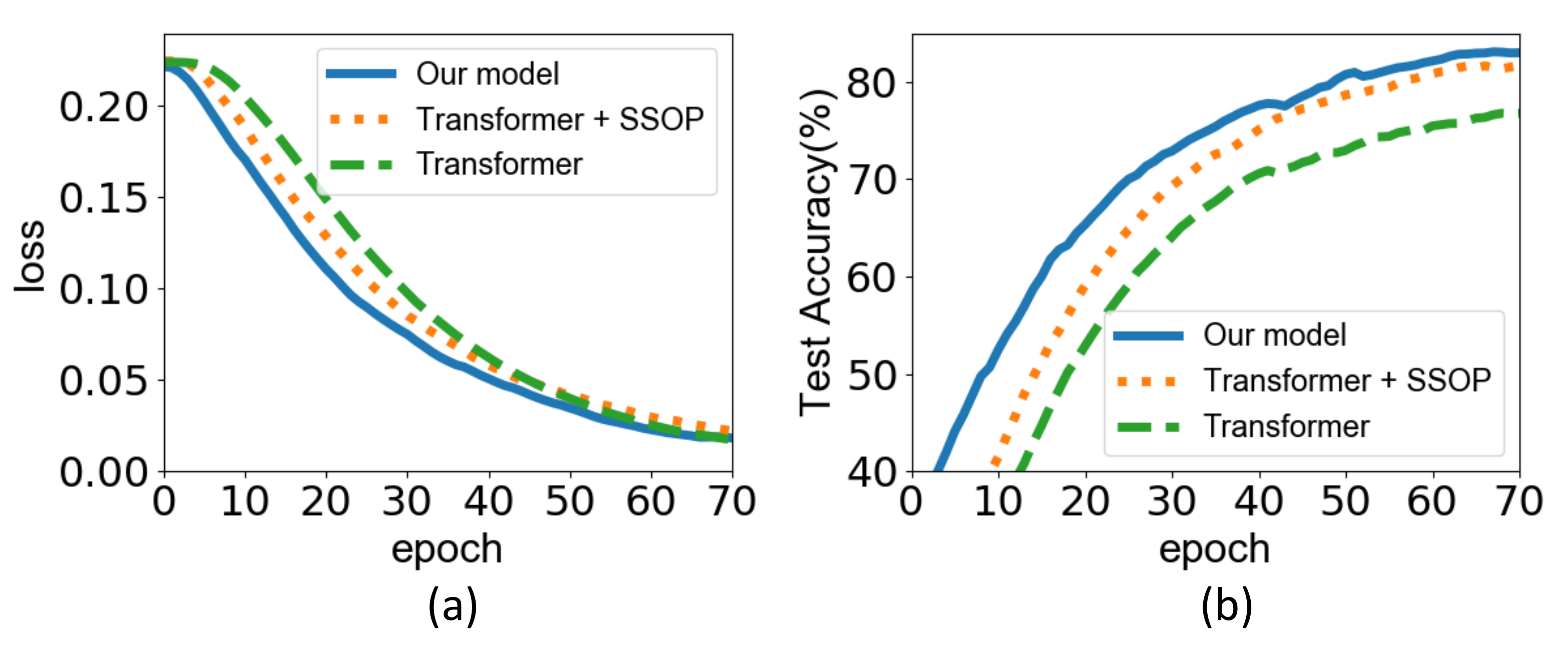}
	\end{center}
	\caption{The learning procedure for the EST with different components during training and testing. (a) The training loss variation in terms of epochs, (b) the testing accuracy variation in terms of epochs.}
	\label{fig:loss_acc}
\end{figure}

Fig.~\ref{fig:loss_acc} shows the training and prediction procedures of the EST with different components. The green dotted curves belong to the baseline Transformer. From the view of the decline rates of training losses, obviously, the gradual addition of AA-SFE and SSOP improved the Transformer performance on both training speed and stability. Meanwhile, the proposed EST with AA-SFE and SSOP is easier to achieve higher accuracy on the test set.

\section{Effect of Important Parameters}
\begin{figure}[htp]
	\begin{center}
		\includegraphics[width=1.0\linewidth]{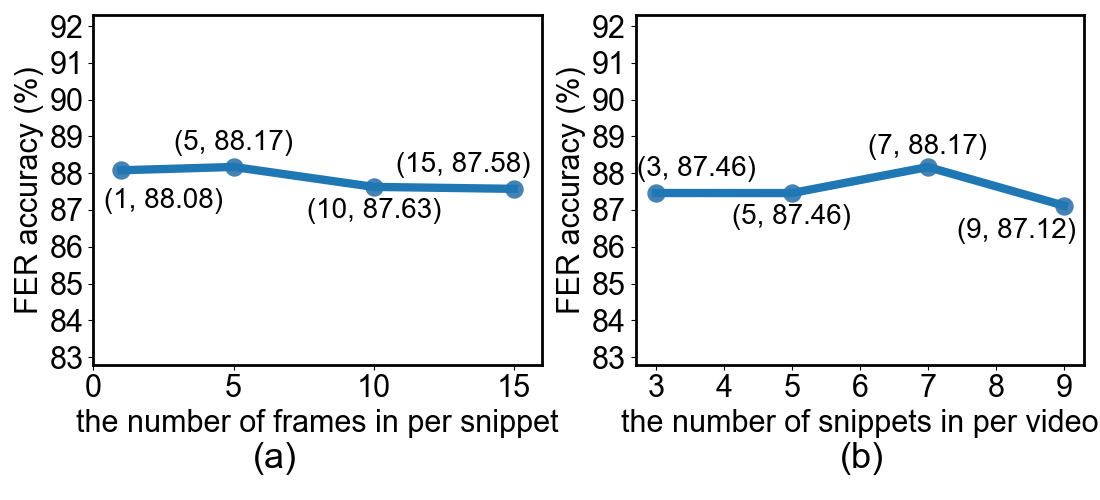}
	\end{center}
	\caption{The impact of the number of frames in per snippet and the number of snippets in per video for FER on the BU-3DFE dataset. (a) The effect of the number of frames in per snippet, (b) the effect of the number of expression snippets in per video.}
	\label{fig:snippet_ablation}
\end{figure}

In Fig.~\ref{fig:snippet_ablation}, we present the FER accuracy curves on the BU-3DFE dataset, which effected by the number of frames in per snippet and the number of snippets in per video. As shown in the Fig.~\ref{fig:snippet_ablation} (a), the accuracy achieved to the highest 88.17$\%$ when we set the number of frames of each snippet to 5. Fig.~\ref{fig:snippet_ablation} (b) shows that the accuracy reached the highest when the number of snippets in per video is set to 7.
Besides, we also observe that different snippet amounts and snippet lengths only resulted in minor performance changes, suggesting that these hyperparameters are less important to our method.
%Hence, in this study, e finally chose to set the number of snippets of each video to 7 and the number of frames of each snippet to 5
Hence, thanks to the long-range relation modeling ability of the Transformer, our EST can be easily extended to adapt to videos of almost any length upon proper training. %We only use 7 snippets because the datasets we used usually contain short videos with around 100 frames.  

\begin{figure}[h]
	\begin{center}
		\includegraphics[width=1.0\linewidth]{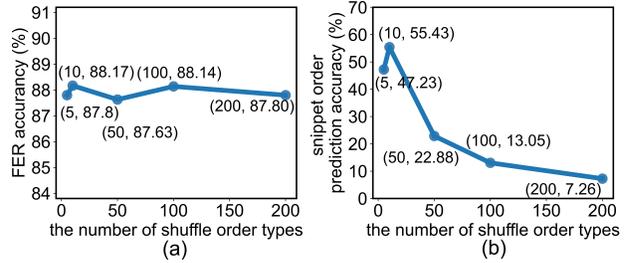}
	\end{center}
	\caption{The influence of the types of shuffle order for FER accuracy and order prediction accuracy on the BU-3DFE dataset. (a) FER accuracy, (b) snippet order prediction accuracy.}
	\label{fig:snippet_frame}
\end{figure}
To evaluate the effect of the types of shuffle order in SSOP learning, Fig.\ref{fig:snippet_frame} presents the variation curves of FER accuracy and snippet order prediction accuracy according to the number of shuffle order types on the BU-3DFE dataset. As shown in Fig.\ref{fig:snippet_frame} (a)(b), when the number of the types is 10, both the FER accuracy and snippet order prediction accuracy reach the highest 88.17$\%$ and 55.43\%, respectively. %And we can see from the Fig.\ref{fig:snippet_frame} (b), the FER accuracy reaches the highest 90.83$\%$ when the types of shuffle order is 5 or 10. Obviously, when the types of shuffle order is set as 10, both the FER precision and reconstruction precision can reach the highest, 
Therefore, during the training, we set the types of shuffle order to 10.

\begin{figure}[htp]
	\begin{center}
		\includegraphics[width=1.0\linewidth]{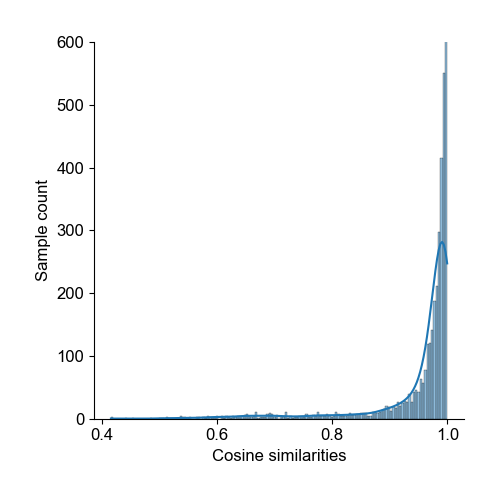}
	\end{center}
	\caption{The distribution of cosine similarity}
	\label{fig:cos_prob}
\end{figure}

Additionally, Fig.\ref{fig:cos_prob} presents the calculation results of cosine similarity in the AA-SFE. %in order to avoid the occurrence of negative operation in the calculation of cosine similarity, all weights were also clipped to an exponential order. Meanwhile, 
Due to small movement variation between expression snippets, all of the calculated cosine similarities are above 0.4, avoiding the occurrence of negative operation in the feature summation. And we set a small value $1e-8$ to avoid division by zero.

\section{Comparison of Different Backbones}
\begin{table}[hpbt]
	\footnotesize
	\centering
	\begin{tabular}{l|cc|c}
		\hline
		Backbone & Params(M) & MACs(G) & Acc(\%)
		\\	
		\hline\hline
		ResNet-18\cite{he2016deep} & \textbf{42.78} & \textbf{63.89} & \textbf{54.26} \\
		ResNet-34\cite{he2016deep} & 52.88 & 128.54 & 51.99 \\
		ResNet-50\cite{he2016deep} & 272.49 & 144.06 & 53.12\\
		%ResNet-101 & xxx & xxx & xxx\\
		\hline
	\end{tabular}
	\caption{Comparison of different backbones.}
	\label{tab:comparison_complexity_efficiency}
\end{table}
Table.~\ref{tab:comparison_complexity_efficiency} reports model parameters,  computational cost, and FER accuracies of three different backbones in processing the AFEW dataset. Obviously, as the backbone network deepens, the proposed model requires more parameters and computational cost. However, the deeper models did not bring a significant increase in FER accuracy. On the contrary, the smaller ResNet-18 as the backbone resulted in the best performance (the FER accuracy of 54.26\%).

\section{Visualization Results of Expression Representations}
% \begin{figure}[htpb]
% 	\begin{center}
% 		%\fbox{\rule{0pt}{2in} \rule{0.9\linewidth}{0pt}}
% 		\includegraphics[width=1.0\linewidth]{fig/T-SNE_feature.png}
% 	\end{center}
% 	\caption{The comparison of different representations in 2D space by t-SNE visualization. (a) The unified salient emotion representation learned by EST, (b) the frame-based features learned by FAN \cite{meng2019frame}, (c) the sequence-based features learned by LSTM, (d) the frame-based features learned by ResNet-18.}
% 	\label{fig:T-SNE_feature}
% \end{figure}

\begin{figure}[hbtp]
	\begin{center}
		\includegraphics[width=1.0\linewidth]{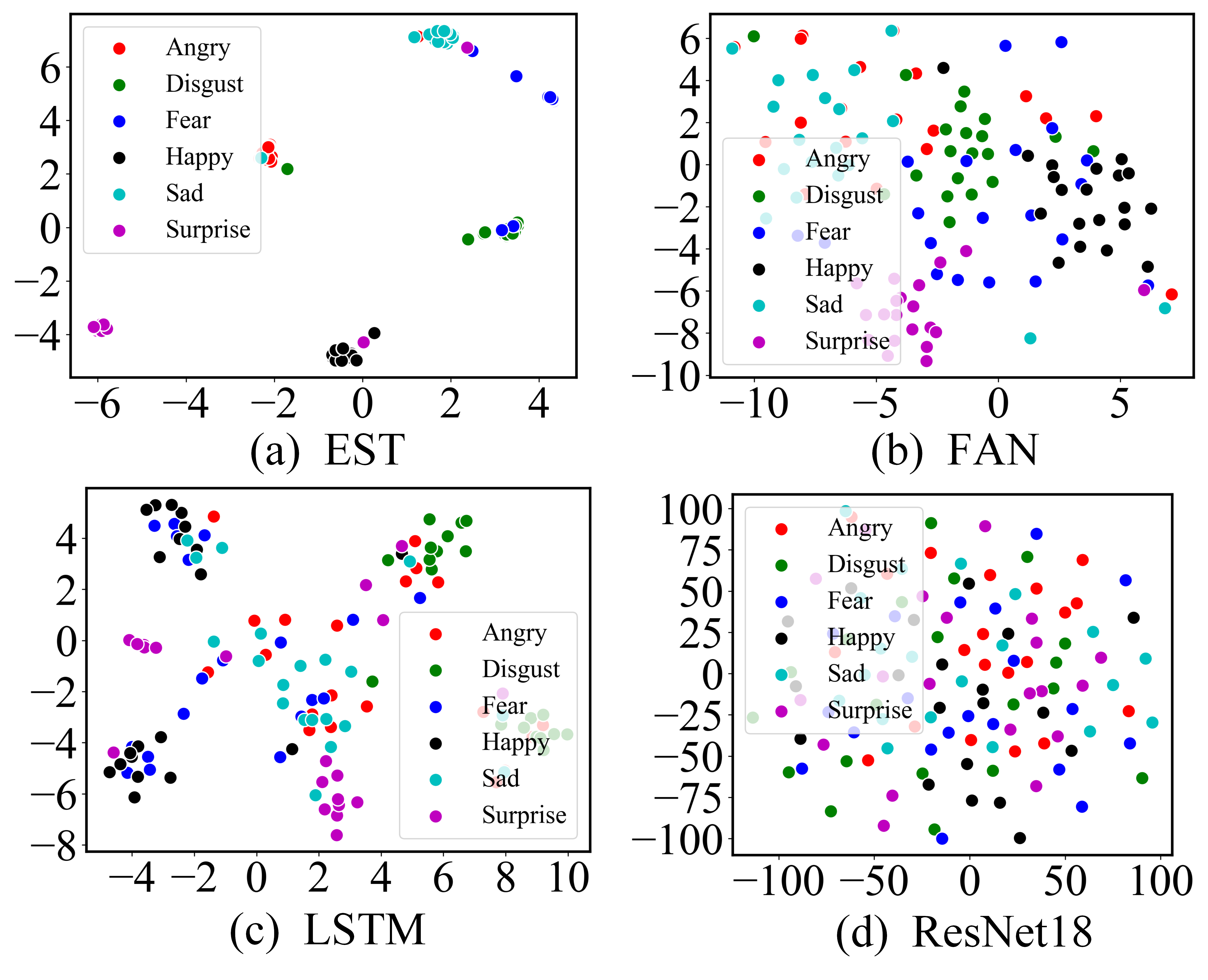}
	\end{center}
	\caption{The comparison of different representations in 2D space by t-SNE visualization. (a) The unified salient emotion representation learned by EST, (b) the frame-based features learned by FAN \cite{meng2019frame}, (c) the sequence-based features learned by LSTM, (d) the frame-based features learned by ResNet-18.}
	\label{fig:T-SNE_feature}
\end{figure}

In Fig.~\ref{fig:T-SNE_feature}, we visualized the emotion features with different settings in a 2D feature space by using the t-SNE~\cite{maaten2008visualizing} on the BU-3DFE dataset. Compared the other emotion features, we can observe that the comprehensive and robust expression representation learned by the EST includes more visual change cues and can significantly be separated according to different expression categories. %It is evident that the proposed ECL-Net can learn more expression-discriminative representations for FER.

\section{Visualization Results of Attention Weights}
\begin{figure}[hpbt]
	\begin{center}
		\includegraphics[width=1.0\linewidth]{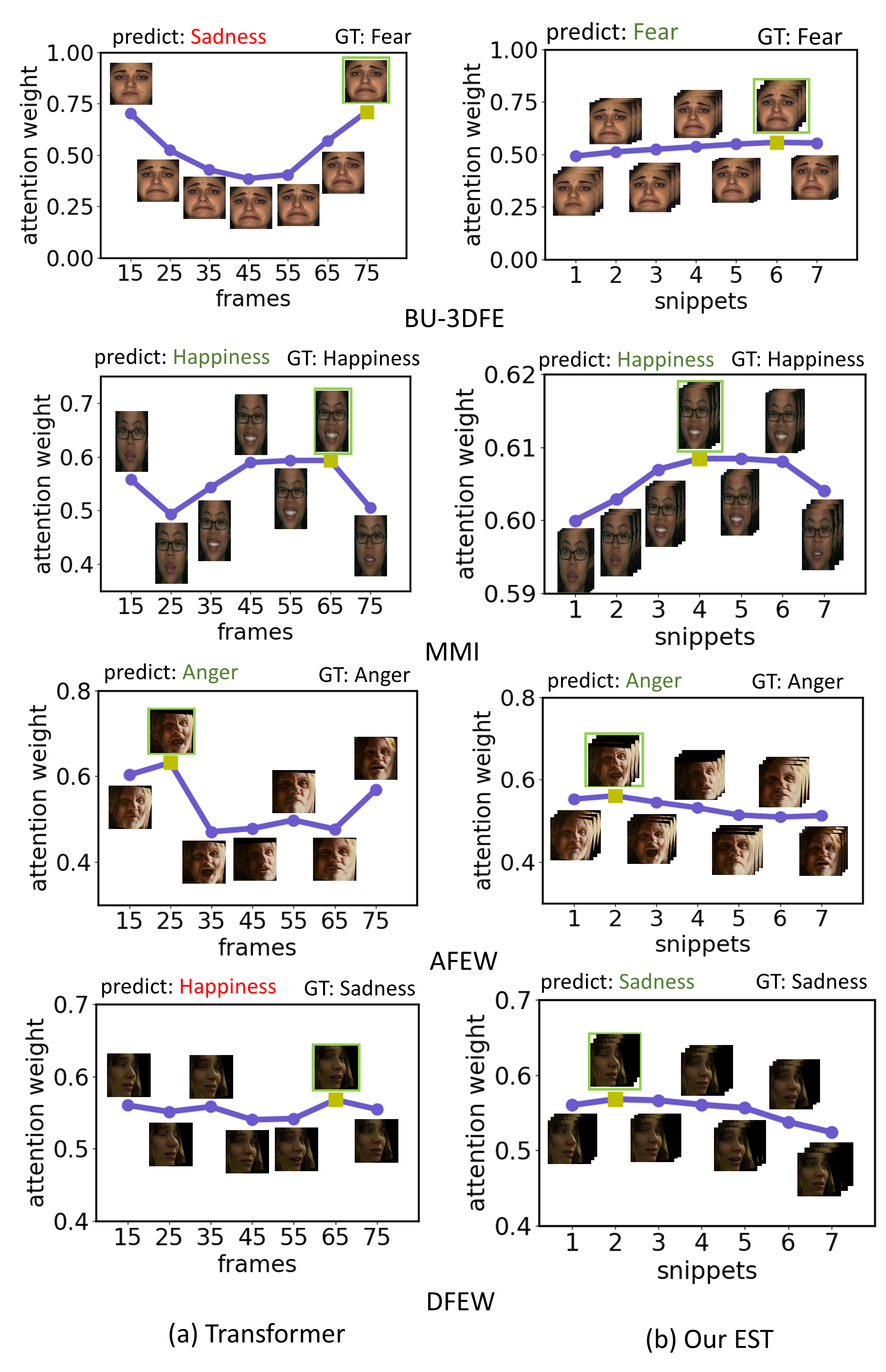}
	\end{center}
	\caption{Comparison of modeling subtle facial expression movements in FER on the BU-3DFE, MMI, AFEW, DFEW dataset. (a) vanilla Transformer, (b) our EST. Note: the green square is located at the position of the most informative expression snippet with the most attention weight.}
	\label{fig:emotion_weight}
\end{figure}

 Fig.~\ref{fig:emotion_weight} shows the comparison of expression relation curves for modeling subtle facial expression movements between the vanilla Transformer and the proposed EST on four videos from the BU-3DFE, MMI, AFEW, DFEW dataset, respectively. %which can describe the emotion saliency weight of each snippet in a video.
From the Fig.~\ref{fig:emotion_weight}(a), the vanilla Transformer tends to focus on the frames with peak expression patterns, which are easily affected by noises such as head pose and other non-expression changes. On the contrary, by decomposing videos into snippets, although the changes of expression movements of intra-/inter-snippets are very subtle in a video, our EST focuses on all expression snippet changes and improves modeling of intra-snippet and inter-snippet subtle facial expression changes, meanwhile, effectively locates the most informative expression snippet by the modelled expression attention weights (see the values of the Ordinate in Fig.~\ref{fig:emotion_weight}(b)).
It is evident that the EST succeeded at focusing on comprehensive expression movements according to the relations of snippet features to the \emph{emotion query embedding} and achieved more robust FER.

\section{Detailed Transformer architecture}
\begin{figure}[thpb]
	\begin{center}
		\includegraphics[width=1.0\linewidth]{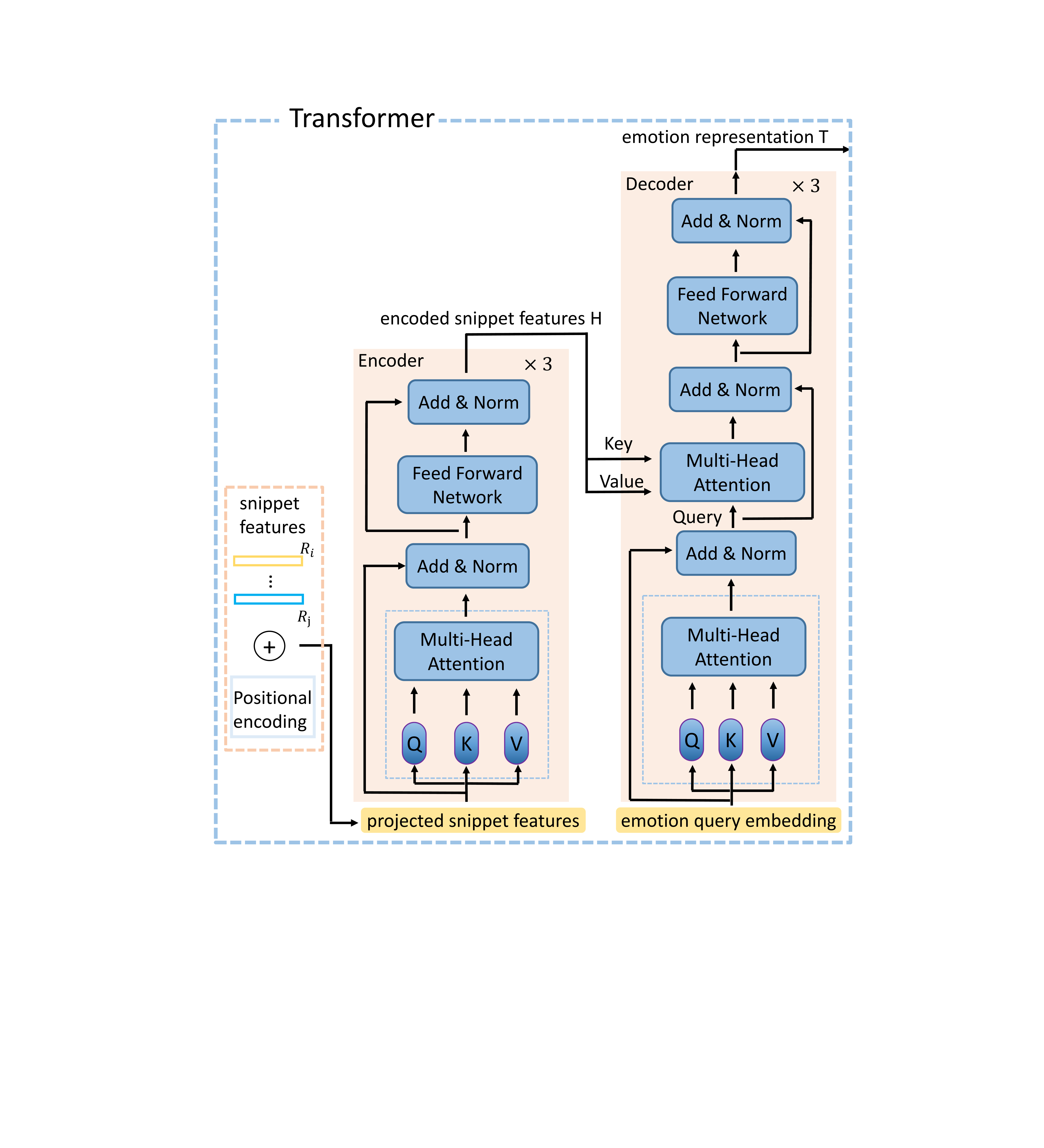}
	\end{center}
	\caption{Architecture of the transformer used in EST.}
	\label{fig:architecture}
\end{figure}

The detailed description of the transformer used in EST, is given in Fig.~\ref{fig:architecture}. Snippet features $\mathcal{R}$ from the AA-SFE extractors are first passed through the transformer encoder, together with positional encoding. Then the decoder receives the emotion query embedding (initially from $\mathcal N(0,1)$) and the encoded snippet features $H$, and produces the emotion representation $T$ through three decoder layers.

{\small
	\bibliographystyle{plain}
	\bibliography{supplement}
}